\documentclass{article}
\PassOptionsToPackage{table}{xcolor}
\usepackage{iclr2027_conference,times}

\usepackage{amsmath,amsfonts,bm}

\def\eqref#1{equation~\ref{#1}}

\def\1{\bm{1}}

\DeclareMathAlphabet{\mathsfit}{\encodingdefault}{\sfdefault}{m}{sl}
\SetMathAlphabet{\mathsfit}{bold}{\encodingdefault}{\sfdefault}{bx}{n}

\usepackage{hyperref}
\usepackage{url}
\usepackage{graphicx}
\usepackage{multirow}
\usepackage{amsmath,amssymb}
\usepackage{listings}
\usepackage{booktabs,tabularx,array,threeparttable}
\usepackage{wrapfig}
\usepackage{capt-of}
\usepackage{longtable}
\usepackage{xltabular}
\usepackage{adjustbox}
\usepackage{caption}
\usepackage[table]{xcolor}
\usepackage{tcolorbox}
\tcbuselibrary{skins,breakable}
\usepackage{tikz}
\usepackage{colortbl}
\usepackage{placeins}

\hypersetup{hidelinks}

\definecolor{completionColor}{HTML}{BBA9CF}
\definecolor{safetyColor}{HTML}{D5A5B0}
\definecolor{behaviorColor}{HTML}{A6CAD8}

\newcommand{\plcicon}[1]{%
  \tikz[baseline=-0.5ex, x=1ex, y=1ex,
        line width=0.10ex, line cap=round, line join=round]{%
    \path[use as bounding box]
      (-0.85,-0.80) rectangle (0.85,0.80);
    #1%
  }%
}

\newcommand{\plcmet}{%
  \plcicon{\draw (0,0) circle[radius=0.62ex];}%
}

\newcommand{\plcpartial}{%
  \plcicon{\draw (-0.72,-0.50) -- (0.72,-0.50)
    -- (0,0.74) -- cycle;}%
}

\newcommand{\plcnotreported}{%
  \plcicon{\draw (-0.55,-0.55) -- (0.55,0.55);
    \draw (-0.55,0.55) -- (0.55,-0.55);}%
}

\graphicspath{{figures/}}

\title{PLCWorld: Benchmarking LLM-Generated PLC Programs in Closed-Loop Plant Simulation}

\author{
Yunji Kim\textsuperscript{1}\thanks{Equal contribution.}\quad
Yunseok Lee\textsuperscript{1}\footnotemark[1]\quad
Hyunwoo Seo\textsuperscript{2}\quad
Jaerim Choi\textsuperscript{2}\quad
Woojin Lee\textsuperscript{1}\thanks{Corresponding author.}\\[0.5em]
\textsuperscript{1}Dongguk University\quad
\textsuperscript{2}MOAI technologies\\[0.3em]
\texttt{\{2022113147,yslee0005,wj926\}@dgu.ac.kr}\\
\texttt{\{yule,jrim\}@moai.co}
}

\newcommand{\taskexample}[1]{``\textit{#1}''}

\iclrfinalcopy
\begin{document}

\maketitle

\fancyhead{}
\fancyfoot{}
\fancyhead[L]{Preprint}
\fancyfoot[C]{\thepage}
\renewcommand{\headrulewidth}{0pt}
\renewcommand{\footrulewidth}{0pt}

\begin{abstract}
  Programmable logic controllers (PLCs) coordinate industrial
  equipment by reading sensor inputs and issuing control commands.
  Evaluating whether large language model (LLM)-generated PLC
  programs satisfy task requirements and safety constraints
  requires observing how their commands affect device and
  workpiece states.
  We introduce \textbf{PLCWorld}, a common closed-loop execution
  environment and benchmark that couples Structured Text (ST)
  execution with simulated plant responses and sensor feedback.
  Grounded in control relations identified in industrial PLC
  programs and engineering documentation, \textbf{PLCWorld} contains
  100 synthetic tasks and 473 registered task--condition pairs
  across \textit{Motion Control} and \textit{Material Handling},
  with difficulty defined by control-dependency scope.
  A common protocol reports \textit{Task Success} and
  \textit{Safety Violation} separately.
  Validation combines practitioner review, reference and alternative programs, targeted counterexamples, specification--evaluator alignment checks, and comparisons
with independent ST runtimes. Reference and alternative
programs satisfy their applicable cases, while all 542
targeted counterexamples activate their designated evaluator
rules under at least one registered condition.
\textit{Execution Gap} relates submission-profile acceptance to subsequent task failure or observed \textit{Safety Violation}.
Across the constructed task groups, direct GPT-5.5 achieves
82.70\% \textit{Task Success} on Easy cases but 25.10\% on Hard
cases. Evaluations of six LLMs and four adapted
generation-and-verification workflows further expose
differences between completion, safety, and generation cost. Our code is available at https://yunji0516.github.io/PLCWorld/.
  \end{abstract}

\section{Introduction}

  A Programmable Logic Controller (PLC) is an industrial controller that coordinates a production process
  by reading sensor signals and controlling devices such as motors, conveyors, and
  grippers~\citep{vyatkin2013software}. PLC control involves dependencies within a basic control function, between
  functions of a job, and between jobs that share equipment. These dependencies arise from device response
  timing, required device and workpiece states, and access to shared resources. They determine when an
  operation may start, which states must be maintained during execution, and when the operation can be
  considered complete. Failing to respect these dependencies can prevent task completion or violate safety
  constraints. Such concerns have long motivated research on the formal verification of PLC
  programs~\citep{ovatman2016overview,adiego2015applying}.

To express such control logic, engineers write programs in the IEC 61131-3 languages~\citep{iec61131}, among which Structured Text (ST) is a high-level textual language. Recent work generates ST code from natural-language requirements with large language models (LLMs), using document and code-example retrieval~\citep{koziolek2026spec2control,yang2025autoplc}, compiler feedback~\citep{fakih2024llm4plc}, and formal verification~\citep{liu2026agents4plc} to support generation and revision. Evaluation has also expanded to project context and execution outcomes~\citep{tu2026semaplc}. The evidence provided by these evaluations depends on the properties, inputs, and execution conditions they cover. We focus on evaluating whether an LLM-generated ST program can coordinate the required operations to complete a plant task while satisfying its safety constraints under specified operating conditions.

On physical equipment, issuing a command does not imply that the corresponding operation has completed. Whether a dependent operation can start or continue depends on the actual state of the controlled equipment and process, collectively called the plant. The plant state evolves as devices respond to commands and interact with workpieces. A PLC program must therefore coordinate its commands with the device and workpiece states required by the task.

\begin{figure}[t]
\centering
\includegraphics[width=0.95\linewidth,height=\textheight,keepaspectratio]{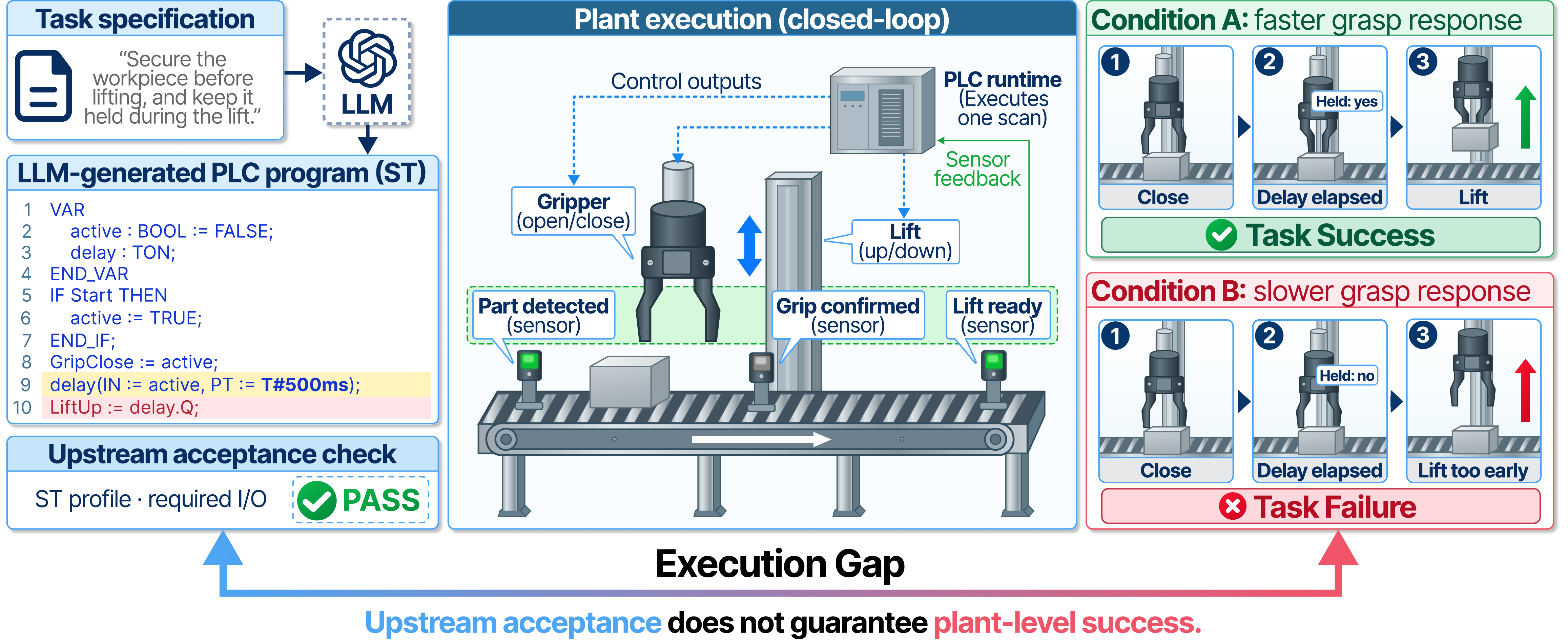}
\caption{\textbf{\textit{Execution Gap}.} Even after the specified upstream checks pass, the lift can rise without the workpiece if grasping is incomplete when the fixed wait ends.}
\vspace{-5pt}
\label{fig:figure1}

\end{figure}

Figure~\ref{fig:figure1} illustrates this dependency with a gripper that must secure a workpiece before lifting it. The generated ST program commands the gripper to close, waits for a fixed interval, and then commands the lift. If grasping takes longer than this interval, even within the specified operating range, the lift starts before the workpiece is secured, leaving it on the table. The failure arises because the program advances to lifting without satisfying the prerequisite grasp state, despite issuing the commands in the intended order.

In plant operation, the outcome of one operation can determine the conditions under which subsequent operations may proceed. Evaluating these dependencies calls for closed-loop plant execution, in which the program's control outputs change device and workpiece states, and sensor inputs derived from those states are returned to the program for subsequent control. An evaluation environment must sustain this interaction throughout the task so that task completion and safety can be assessed from the resulting plant behavior.

We introduce \textbf{PLCWorld}, an execution environment and benchmark for evaluating LLM-generated ST programs on industrial control tasks. Our contributions are as follows. (1) We develop a closed-loop plant execution environment that couples ST execution with simulated device and workpiece behavior at each PLC scan. (2) We construct \textit{Motion Control} and \textit{Material Handling} task suites grounded in control relations identified in industrial PLC programs and engineering materials, with explicit task objectives, control dependencies, operating conditions, and safety constraints. (3) We establish a common evaluation protocol that reports \textit{Task Success} and \textit{Safety Violation} separately and compares upstream checks with plant execution outcomes on the same submissions to analyze \textit{Execution Gap}.

We evaluate six LLMs and four generation-and-verification methods on
PLCWorld. Across both settings, Hard tasks show substantially lower
\textit{Task Success} rates than Easy and Medium tasks, while
\textit{Safety Violation} captures failures not reflected by
completion alone. The evaluated
    adaptations show mixed completion and safety
changes versus matched direct generation, without consistent
plant-level gains. Substantial failures persist after common
upstream acceptance. Failure analysis identifies failure to
establish valid completion by the deadline as the most frequent
recorded signature.

\section{Related Work}

\textbf{LLM-based PLC code generation.} Recent work has explored LLM-based generation and validation of PLC control code using retrieval, compiler feedback, formal verification, expert review, and runtime testing. Representative approaches include LLM4PLC~\citep{fakih2024llm4plc}, Agents4PLC~\citep{liu2026agents4plc}, AutoPLC~\citep{yang2025autoplc}, Spec2Control~\citep{koziolek2026spec2control}, and SemaPLC~\citep{tu2026semaplc}. Table~\ref{tab:related_scope} summarizes the validation scope reported by these studies.

\begin{table}[t]
\centering
\begin{threeparttable}
\caption{Reported validation coverage of LLM-based PLC code generation studies.}
\label{tab:related_scope}
\footnotesize
\setlength{\tabcolsep}{3pt}
\renewcommand{\arraystretch}{1.15}
\begin{tabularx}{0.95\linewidth}{@{}l*{6}{>{\centering\arraybackslash}X}@{}}
\toprule
& \multicolumn{2}{c}{\textbf{Program validation}}
& \multicolumn{2}{c}{\textbf{Dependency validation}}
& \multicolumn{2}{c}{\textbf{Plant outcomes}} \\
\cmidrule(lr){2-3}\cmidrule(lr){4-5}\cmidrule(l){6-7}
\textbf{Study}
& Syntax /\newline build
& Program\newline execution
& Control\newline dependencies
& Workpiece\newline handling
& Task\newline completion
& Safety\newline violations \\
\midrule
LLM4PLC
& \plcmet & \plcmet & \plcmet
& \plcpartial & \plcpartial & \plcnotreported \\
Agents4PLC
& \plcmet & \plcmet & \plcmet
& \plcnotreported & \plcnotreported & \plcnotreported \\
AutoPLC
& \plcmet & \plcmet & \plcmet
& \plcnotreported & \plcnotreported & \plcnotreported \\
Spec2Control
& \plcpartial & \plcnotreported & \plcpartial
& \plcnotreported & \plcnotreported & \plcnotreported \\
SemaPLC
& \plcmet & \plcmet & \plcmet
& \plcnotreported & \plcnotreported & \plcnotreported \\
\midrule
\textbf{PLCWorld}
& \plcmet & \plcmet & \plcmet & \plcmet & \plcmet & \plcmet \\
\bottomrule
\end{tabularx}
\begin{tablenotes}[flushleft]
\footnotesize
\setlength{\itemsep}{1pt}
\item[] \plcmet: assessed; \plcpartial: partially assessed;
\plcnotreported: not reported.
\end{tablenotes}
\end{threeparttable}
\end{table}

Prior studies assess control dependencies through formal verification and runtime tests, with LLM4PLC also reporting a physical handling demonstration. The plant-outcome columns concern task completion and safety judgments derived from physical or simulated plant states and events, beyond agreement between program outputs and reference traces. PLCWorld brings device and workpiece dependencies into a common closed-loop execution environment and evaluates \textit{Task Success} and \textit{Safety Violation} separately under specified operating conditions.

\textbf{PLC development and commissioning tools.} Existing PLC tools support controller simulation~\citep{mitsu_gx_sim}, virtual commissioning with equipment models~\citep{mitsu_gemini,siemens_virtual_commissioning}, and execution of IEC 61131-3 programs through open runtimes~\citep{alves2018openplc,openplc_runtime}. Comparing LLM-generated programs using these tools requires common tasks, operating conditions, and criteria for task completion and safety. PLCWorld combines a closed-loop plant execution environment with task suites organized by control dependencies and a common evaluation protocol, enabling systematic comparisons across models and generation-and-verification methods.

\textbf{Execution-based evaluation.} Recent work evaluates LLM outputs through their consequences in executable environments. BIRD-INTERACT~\citep{huo2026bird} evaluates interactive database tasks, BuildArena~\citep{xia2025buildarena} tests generated engineering structures in physics simulation, and MCCoder~\citep{li2025mccoder} verifies generated Python motion-control code in simulation. For PLC programs, such evaluation must preserve the control dependencies through which outputs change device and workpiece states and those states, in turn, constrain subsequent control.

\section{PLCWorld}

\begin{figure}[h]
\centering
\includegraphics[width=0.95\linewidth,keepaspectratio]{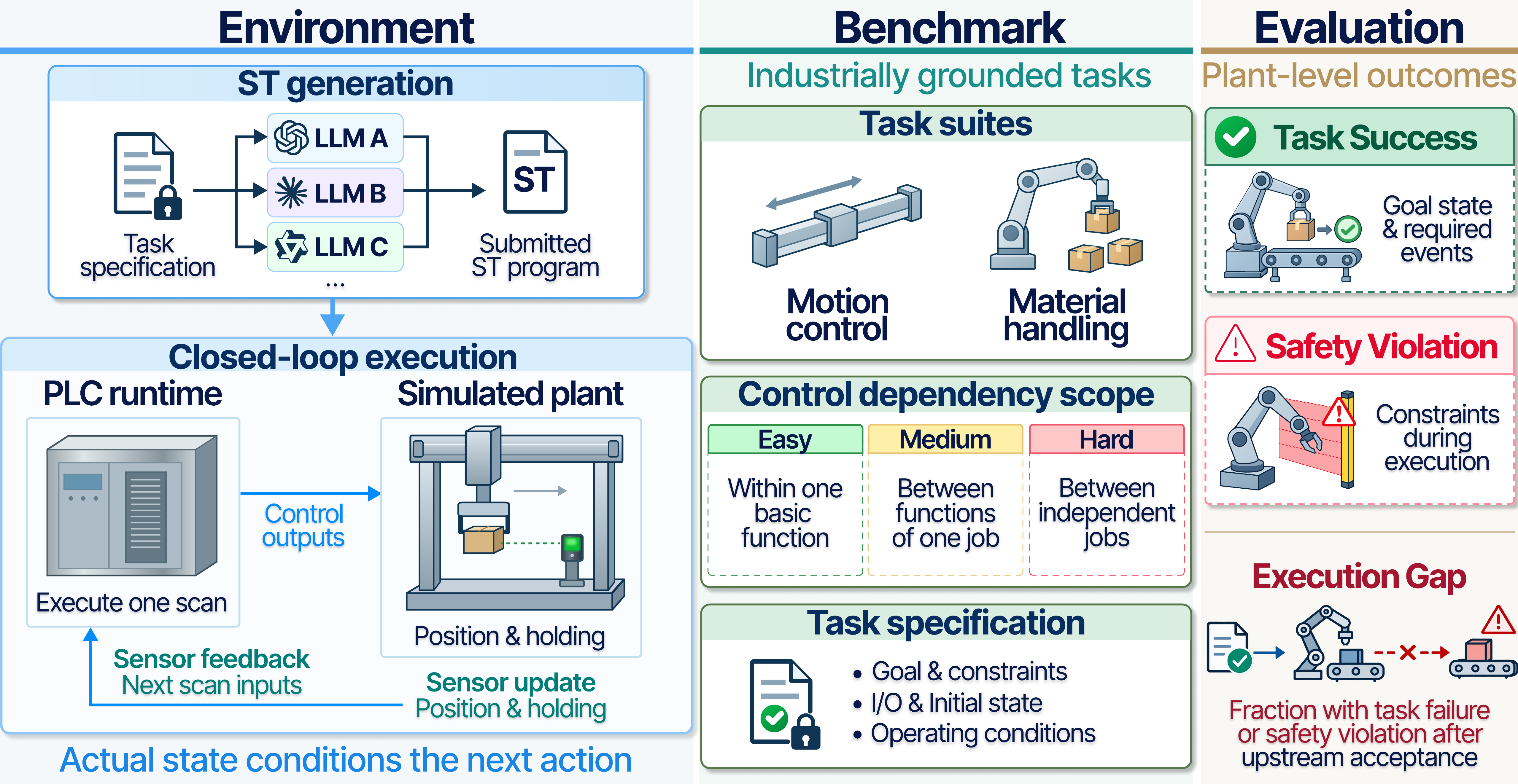}
\caption{\textbf{The PLCWorld framework.} ST programs generated from task specifications are executed in closed loop with a simulated plant. Execution records are used to assess \textit{Task Success} and \textit{Safety Violation} separately, and comparison with upstream checks on the same programs reveals \textit{Execution Gap}.}
\label{fig:figure2}
\end{figure}

\textbf{PLCWorld} combines a plant execution environment, benchmark task suites, and an evaluation protocol for LLM-generated ST programs (Figure~\ref{fig:figure2}). Task specifications define the required outcomes and operating conditions, and submitted programs are evaluated through interaction with a simulated plant. The resulting execution records are used to assess \textit{Task Success} and \textit{Safety Violation} separately.

\subsection{Plant Execution Environment}

PLC operations depend on device and workpiece states that determine when they may start, continue, or complete. Issuing a command does not necessarily establish the required state, so evaluating these dependencies requires following the command's effects through plant execution. PLCWorld couples each generated ST program to a simulated plant that updates device and workpiece states and returns sensor inputs derived from those states for subsequent control.

\textbf{Closed-loop execution.} Each execution starts from an ST program, a device configuration with I/O bindings, an initial plant state, and execution conditions. The ST program runs cyclically in PLC scans, reading sensor inputs and computing control outputs while retaining its internal state across scans. After each scan, the plant advances under these outputs, and sensor observations of its updated state provide inputs for the next scan. Equation~(\ref{eq:loop}) expresses this cycle over a scan interval \(\Delta t\), where \(X_t\), \(Y_t\), \(C_t\), and \(P_t\) denote sensor inputs, control outputs, retained program state, and plant state; \(\theta_t\) specifies external execution conditions, \(F\) updates the plant state, and \(H\) derives sensor observations.

\begin{equation}
\label{eq:loop}
(C_{t+1},Y_t)=\operatorname{Exec}(C_t,X_t;\Delta t),\quad
P_{t+1}=F(P_t,Y_t,\theta_t,\Delta t),\quad
X_{t+1}=H(P_{t+1},\theta_{t+1}).
\end{equation}

Timers advance by the scan interval \(\Delta t\). External inputs, such as start signals and job requests, are specified by the execution conditions, whereas sensor feedback is derived from the updated plant state rather than replayed from a prerecorded trace. The supported ST profile specifies the supported ST syntax, function-block rules, and initialization.

\textbf{Plant state.} The plant state \(P_t\) represents device positions and operating states, workpiece locations, and holding and support relations. An acquisition or handover takes effect only when the device state and workpiece location satisfy the required conditions. For example, if a workpiece is outside the gripper's acquisition region, closing the gripper establishes no holding relation, and subsequent motion leaves the workpiece behind. The simulator models device operation and response times, sensor feedback, and the transitions for acquisition, holding, and handover, but not contact forces or mechanical deformation.

\textbf{Execution records.} Execution runs to the evaluation endpoint or termination condition regardless of completion claims. At each scan, PLCWorld records sensor inputs \(X_t\), control outputs \(Y_t\), the updated plant state \(P_{t+1}\), and events. Program errors are recorded with their interruption points. These records support the separate assessment of \textit{Task Success} and \textit{Safety Violation}, failure analysis, and 3D replay.

\begin{figure}[h]
\centering
\includegraphics[width=\linewidth,keepaspectratio]{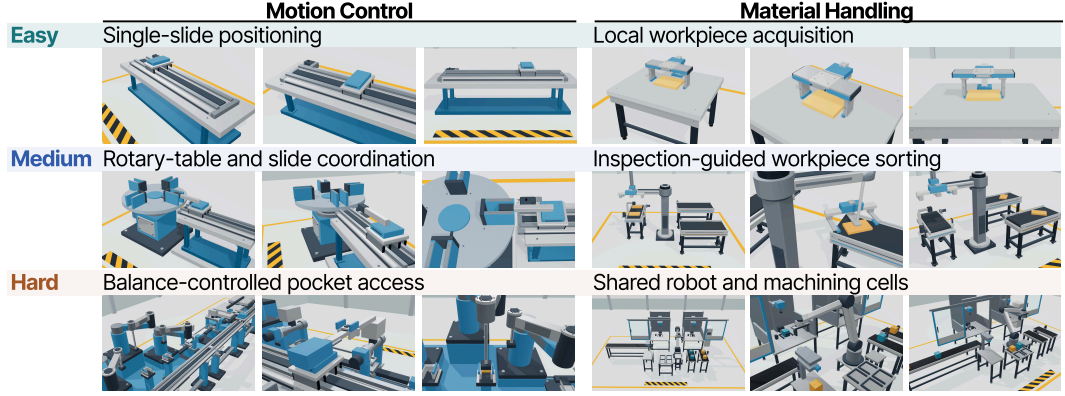}
\caption{\textbf{Representative executions by suite and difficulty.} Columns show \textit{Motion Control} (left) and \textit{Material Handling} (right), and rows show control dependencies within one basic function (Easy), across the functions of one job (Medium), and between jobs sharing resources (Hard).}
\label{fig:figure3}
\end{figure}

\begin{table*}[h]
\centering
\begin{threeparttable}
\caption{Difficulty levels defined by dependency scope, with tasks and representative examples.}
\label{tab:difficulty_levels}

\footnotesize
\setlength{\tabcolsep}{5pt}
\renewcommand{\arraystretch}{1.12}

\renewcommand{\tabularxcolumn}[1]{m{#1}}
\providecommand{\taskexample}[1]{#1}

\begin{tabularx}{0.98\linewidth}{
    @{}l c
    >{\hsize=0.9\hsize\linewidth=\hsize
      \raggedright\arraybackslash}X
    >{\hsize=1.05\hsize\linewidth=\hsize
      \raggedright\arraybackslash}X
    >{\hsize=1.05\hsize\linewidth=\hsize
      \raggedright\arraybackslash}X@{}
}
\toprule
\textbf{Level}
& \textbf{Tasks\tnote{a}}
& \textbf{Dependency Scope}
& \textbf{\textit{Motion Control}}
& \textbf{\textit{Material Handling}} \\
\midrule

Easy
& 15
& Within one basic function
& \taskexample{Move a slide to a position, confirm arrival and standstill}
& \taskexample{Grasp a workpiece at a prepared position, confirm it is held} \\
\addlinespace[3pt]

Medium
& 20
& Between functions of one job: one function's state conditions another's execution
& \taskexample{Retract the slide, reorient the table, approach after standstill}
& \taskexample{Transport a workpiece, release it after the jig is ready} \\
\addlinespace[3pt]

Hard
& 15
& Between independent jobs: one job's state constrains another's progress
& \taskexample{Coordinate two lift jobs as the shared shuttle and rotary/approach cell become available}
& \taskexample{Coordinate two handling jobs as the shared robot and inspection station become available} \\

\bottomrule
\end{tabularx}

\begin{tablenotes}[flushleft]
\footnotesize
\setlength{\itemsep}{0pt}
\item[a] Per suite. PLCWorld contains 100 tasks in total.
\end{tablenotes}

\end{threeparttable}
\end{table*}

\subsection{Benchmark Tasks}
\label{sec:benchmark_tasks}

\textbf{Task specification.} Each task specifies a goal plant state, required intermediate events, and their precedence relations. Safety constraints are stated separately. Any implementation satisfying these requirements is valid, regardless of its internal states, branching, or timers.

The model receives these requirements together with device relationships, I/O semantics, the initial plant state, and permitted operating conditions. The specification explains device responses to output activation and deactivation and sensor indications of position, operation completion, and workpiece holding. Evaluation conditions may vary the initial state, device responses, or external inputs within the specified operating range. The specification distinguishes the completion deadline from the evaluation endpoint or termination condition and states any requirement to maintain the completed state.


\textbf{Task suites.} Drawing on established work in \textit{Motion Control}
  \citep{ruderman2020motion} and PLC-based \textit{Material Handling}
  \citep{bassily2007mechatronics}, we organize tasks into two suites by their primary functional outcome. (1) \textit{Motion Control} tasks control device positions and operating states, such as positioning, homing, and coordinated movement. (2) \textit{Material Handling} tasks change workpiece locations and holding relations, such as acquisition, transport, placement, and handover. For example, a transport device that arrives without its workpiece fails the objective.


\textbf{Difficulty levels.} We define difficulty by the scope of control dependencies that a program must coordinate, using the same criterion for both task suites. These levels distinguish coordination within a basic function, between functions, and between jobs, capturing differences in control requirements that device count or code length alone does not describe. The basic unit is a control operation with its own objective and completion condition. A basic function includes its sensor checks, short sequences, and retained state, with boundaries fixed during task construction. Table~\ref{tab:difficulty_levels} defines the levels and provides task counts and examples, while Figure~\ref{fig:figure3} illustrates representative executions. Short sequences or state memory confined to one basic function remain Easy, and serial or fixed-priority control remains valid for Hard tasks if it meets the requirements.

\subsection{Evaluation}
\label{sec:evaluation}

PLCWorld evaluates functional completion (\textit{Task Success})
separately from observed breaches of safety constraints
(\textit{Safety Violation}). A case is an assigned evaluation
of a task's submitted program under one registered execution condition.
For each independent repetition, one submitted ST program per task is held fixed and
evaluated across all its registered conditions, yielding 473 cases for 100 tasks.
A case may succeed despite a violation or fail without one.

\textbf{\textit{Task Success}.}
We assign \(U_e=1\) when the plant reaches the required goal state within the deadline while satisfying all required events, precedence relations, and hold conditions. When a completion signal is required, the plant state after applying the signal for one scan must satisfy
the completion conditions. \textit{Success Rate} (SR) in Eq.~(\ref{eq:rates}) averages \(U_e\in\{0,1\}\) over \(\mathcal{E}\), the set of evaluation cases excluding verified
infrastructure errors.

\textbf{\textit{Safety Violation}.} An execution incurs a violation if it breaches a specified constraint during evaluation, such as conflicting outputs, operation without an interlock, or prohibited movement. Monitoring continues past completion claims, and recorded violations persist. \textit{Safety Violation Rate} (SVR) in Eq.~(\ref{eq:rates}) averages the indicator \(V_e\in\{0,1\}\) over \(\mathcal{E}_{\mathrm{exec}}\subseteq\mathcal{E}\), cases entering plant execution. Violations before interruption count, and observing none does not establish safety over the full window.

\begin{equation}
\label{eq:rates}
\operatorname{SR}=\frac{1}{|\mathcal{E}|}\sum_{e\in\mathcal{E}}U_e,
\qquad\qquad
\operatorname{SVR}=\frac{1}{|\mathcal{E}_{\mathrm{exec}}|}\sum_{e\in\mathcal{E}_{\mathrm{exec}}}V_e.
\end{equation}

Missing or unusable outputs, submission-profile violations, parsing failures, and execution errors attributable to the model count as Task Success failures. Non-executed cases are excluded from SVR, while infrastructure errors are excluded from all metrics and reported separately.

\textbf{\textit{Execution Gap}.}
For an upstream check \(k\), \(\mathrm{ExecutionGap}_k\)
in Eq.~(\ref{eq:gap}) is the fraction of included cases \(\mathcal{A}_k\subseteq\mathcal{E}\) assigned to programs passing \(k\) that show task failure \(\mathcal{F}=\{e\in\mathcal{E}:U_e=0\}\) or an observed violation \(\mathcal{W}=\{e\in\mathcal{E}_{\mathrm{exec}}:V_e=1\}\). Cases with both count once, each check is computed separately, and each check and the plant evaluation use the same submitted program. 

\begin{equation}
\label{eq:gap}
\operatorname{ExecutionGap}_k=\frac{|\mathcal{A}_k\cap(\mathcal{F}\cup\mathcal{W})|}{|\mathcal{A}_k|}.
\end{equation}

\section{Benchmark Construction and Validation}
\label{sec:construction-en}

\subsection{Industrially Grounded Task Construction}

\begin{wrapfigure}{r}{0.42\linewidth}
\vspace{-18pt}
\centering

\includegraphics[width=0.95\linewidth]{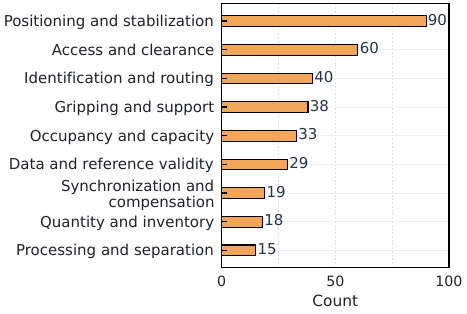}

\captionsetup{
  width=\linewidth,
  singlelinecheck=false,
}

\caption{Control requirement families}

\vspace{-20pt}

\label{fig:control_relations}

\end{wrapfigure}

PLCWorld grounds its task construction in PLC programs and
engineering documentation from real-world manufacturing equipment.
The source archive contains 2,201 PLC and related automation project
files, together with I/O lists, electrical drawings, equipment
manuals, and other engineering documents. We examine these
programs and documents to identify recurring control relations,
including feedback-conditioned actuation, access and clearance
interlocks, support and handover relations, dependencies involving
workpiece states, and coordination through shared equipment.

Each control relation records an action, its observable state or
feedback, the device or workpiece state required before the
action may proceed, and any state that must remain valid during
subsequent operations. We instantiate these relations as synthetic
tasks in a common execution environment, defining task goals,
operating conditions, device arrangements, I/O mappings, and
numerical parameters without reproducing the source implementation.
Figure~\ref{fig:control_relations} summarizes the requirement
families represented in these tasks. These families characterize
task content rather than difficulty, which is defined by dependency
scope. Their suite--difficulty breakdown is reported in
Appendix~\ref{app:task_statistics}.

\subsection{Benchmark Validation}
\label{sec:benchmark_validation}

\begin{wraptable}{r}{0.42\linewidth}
\vspace{-17pt}
\centering
\caption{Overview of validation}
\label{tab:validation_overview}

\scriptsize
\setlength{\tabcolsep}{5pt}
\renewcommand{\arraystretch}{1.0}

\begin{tabularx}{\linewidth}{
  @{}
  >{\raggedright\arraybackslash}p{0.5\linewidth}
  >{\raggedright\arraybackslash}X
  @{}
}
\toprule
\textbf{Validation target}
& \textbf{Validation method} \\
\midrule
\textbf{(1)} Task specification
& Practitioner review \newline \\

\textbf{(2)} Evaluator judgment
& Reference/alternative ST \newline
  \& counterexamples \\

\textbf{(3)} Spec--evaluator alignment
& Binding \& \newline semantic checks \\

\textbf{(4)} ST runtime
& Reruns \& \newline
  cross-runtime checks \\
\bottomrule
\end{tabularx}
\vspace{-15pt}
\end{wraptable}

Reliable evaluation requires plausible, sufficiently specified
tasks, evaluator judgments consistent with
requirements, and reproducible execution with consistent ST semantics. Task feasibility alone does not establish that the evaluator accepts alternative valid implementations or detects violations. As summarized in Table~\ref{tab:validation_overview}, we examine
these requirements through four complementary validation checks, each addressing a distinct aspect of benchmark reliability. 
Detailed procedures and results appear in Appendix~\ref{app:benchmark_validation}.

\paragraph{(1) Task specification validation.}
Practitioner review assesses whether the synthetic tasks are
industrially plausible and sufficiently specified for implementation.
Four practitioners experienced in PLC control and equipment
operation review all 100 task specifications, providing assessments
for the six suite--difficulty groups and recording individual task
issues. They examine information sufficiency, completion and
safety clarity, and dependency-based difficulty levels. Their
feedback informs revisions before specifications are frozen
for evaluation.

\paragraph{(2) Evaluator judgment validation.}
Reference ST programs written to satisfy the task specifications
achieve \textit{Task Success} without an observed
\textit{Safety Violation} across all 100 tasks and 473 registered
task--condition cases. Alternative implementations for a subset
of tasks yield the same outcomes in their tested executions,
supporting acceptance beyond a single solution. Successful
executions alone do not establish violation detection, so we also
test controlled counterexamples targeting completion, precedence,
retention, interlock, command, and safety requirements. Each triggers its prespecified evaluator rule in at least one registered execution condition.

\paragraph{(3) Specification--evaluator alignment validation.}
The evaluator should enforce published task requirements
without introducing unstated conditions. A bidirectional audit
checks whether requirements are represented in executable rules
and whether those rules are supported by the specifications.
Automated checks verify task-input integrity and
consistency of I/O interfaces, scan periods, evaluator
bindings, and submission constraints. Audit findings guide
revisions and regression checks.

\paragraph{(4) ST runtime validation.}
Reference-case reruns reproduce trace hashes and judgments.
Diagnostics use MATIEC/GCC~\citep{beremiz2017matiec};
S7-PLCSIM Advanced~\citep{siemens2024plcsimapi} tests examine
on-delay timer (\texttt{TON}) scan-boundary behavior.
MATIEC/GCC comparisons cover 268 unchanged programs from
six direct-generation models across all registered conditions
with the frozen plant and evaluator.
Traces agree in \(1{,}162/1{,}199\) cases,
\textit{Task Success} in \(1{,}195\), and
\textit{Safety Violation} in all.
Four \textit{Task Success} differences involve one program
calling a \texttt{TON} instance twice per scan.
\newcommand{\plctbd}{--}
\newcommand{\plcna}{N/A}

\section{Experiments}
\label{sec:experiments-en}

\subsection{Experimental Setup}
\label{sec:experimental_setup}

We evaluate two settings.
\textbf{(1) Direct ST generation} compares GPT-5.5~\citep{openai2026gpt55},
GPT-4o~\citep{hurst2024gpt}, Claude Sonnet 5~\citep{anthropic2026sonnet5},
Gemini 3.1 Pro~\citep{google2026gemini31pro},
Gemini 3.8 Flash~\citep{google2026gemini38flash}, and
Qwen3.5-Plus~\citep{qwen3.5}. Each produces three independent
single-turn generations per task without tools, execution feedback,
revision, or best-of selection.
\textbf{(2) Generation-and-verification} compares PLCWorld adaptations
of LLM4PLC~\citep{fakih2024llm4plc}, Agents4PLC~\citep{liu2026agents4plc},
AutoPLC~\citep{yang2025autoplc}, and SemaPLC~\citep{tu2026semaplc}
with GPT-5.5 as the shared backbone
(Appendix~\ref{app:framework_adaptations}).

\begin{wrapfigure}{r}{0.42\columnwidth}
\centering
\vspace{0pt}
\includegraphics[width=\linewidth]{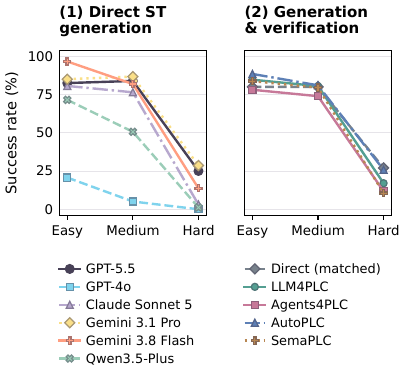}
\caption{Suite-combined success rate by difficulty.}
\label{fig:sr_by_difficulty}

\vspace{15pt} %

\captionsetup{width=\linewidth}
\captionof{table}{Generation cost on the 18-task subset.}
\label{tab:generation_efficiency}

\scriptsize
\setlength{\tabcolsep}{2.2pt}
\renewcommand{\arraystretch}{1.04}

\begin{tabular*}{\linewidth}{@{\extracolsep{\fill}}lrrrr@{}}
\toprule
\textbf{Model / Method} & \textbf{Calls} & \textbf{Revised} & \textbf{Tok. (k)} & \textbf{Time (s)} \\
\midrule
\rowcolor{gray!15}
\multicolumn{5}{l}{\textbf{\textit{(1) Direct ST generation}}} \\
GPT-5.5 & 1 & 0/18 & 11.51 & 99.29 \\
GPT-4o & 1 & 0/18 & 4.45 & 5.90 \\
Claude Sonnet 5 & 1 & 0/18 & 17.00 & 122.32 \\
Gemini 3.1 Pro & 1 & 0/18 & 19.17 & 118.09 \\
Gemini 3.8 Flash & 1 & 0/18 & 25.80 & 65.54 \\
Qwen3.5-Plus & 1 & 0/18 & 19.80 & 250.76 \\
\midrule
\rowcolor{gray!15}
\multicolumn{5}{l}{\textbf{\textit{(2) Generation-and-verification frameworks}}} \\
LLM4PLC   & 4  & 5/18  & 48.28  & 321.13 \\
  Agents4PLC& 5  & 10/18 & 73.11  & 452.56 \\
  AutoPLC   & 4  & 2/18  & 30.96  & 134.34 \\
  SemaPLC   & 18 & 10/18 & 523.88 & 230.90 \\
\bottomrule
\end{tabular*}

\vspace{2pt}
\parbox{\linewidth}{
\scriptsize\raggedright
Calls, tokens, and time are per-task medians. Revised is the
  number of tasks with at least one revision (out of 18).}
\vspace{-55pt} %
\end{wrapfigure}

Each final program is evaluated under all registered task
conditions at the task-specific PLC scan period.
The common upstream check assesses final submissions against
the supported ST profile and required I/O interface, separately
from framework-internal compilation and verification.
SR, SVR, and ExecutionGap follow
Eqs.~(\ref{eq:rates}) and~(\ref{eq:gap}).
We measure generation cost on a fixed, shared 18-task subset
balanced across the six suite--difficulty groups.
Details appear in Appendix~\ref{app:experimental_details}.

\begin{table*}[t]
\centering
\scriptsize
\setlength{\tabcolsep}{2.6pt}
\renewcommand{\arraystretch}{1.08}

\caption{Results (\%) by suite and difficulty.
\textit{Execution Gap} is pooled across suites, conditional
on common upstream acceptance. Bold marks each block's
highest SR and lowest SVR, including ties.}
\label{tab:main_results}

\begin{tabular*}{0.98\textwidth}{
@{\extracolsep{\fill}}
l*{13}{c}
@{}
}
\toprule

\multirow{3}{*}{\textbf{Model / Method}}
& \multicolumn{6}{c}{\textbf{Motion Control}}
& \multicolumn{6}{c}{\textbf{Material Handling}}
& \multirow{3}{*}{
  \shortstack{\textbf{Execution}\\[-1pt]\textbf{Gap$\downarrow$}}
} \\

\cmidrule(lr){2-7}
\cmidrule(lr){8-13}

& \multicolumn{2}{c}{\textbf{Easy}}
& \multicolumn{2}{c}{\textbf{Medium}}
& \multicolumn{2}{c}{\textbf{Hard}}
& \multicolumn{2}{c}{\textbf{Easy}}
& \multicolumn{2}{c}{\textbf{Medium}}
& \multicolumn{2}{c}{\textbf{Hard}}
& \\

\cmidrule(lr){2-3}
\cmidrule(lr){4-5}
\cmidrule(lr){6-7}
\cmidrule(lr){8-9}
\cmidrule(lr){10-11}
\cmidrule(lr){12-13}

& \textbf{SR$\uparrow$}
& \textbf{SVR$\downarrow$}
& \textbf{SR$\uparrow$}
& \textbf{SVR$\downarrow$}
& \textbf{SR$\uparrow$}
& \textbf{SVR$\downarrow$}
& \textbf{SR$\uparrow$}
& \textbf{SVR$\downarrow$}
& \textbf{SR$\uparrow$}
& \textbf{SVR$\downarrow$}
& \textbf{SR$\uparrow$}
& \textbf{SVR$\downarrow$}
& \\

\midrule

\rowcolor{gray!15}
\multicolumn{14}{l}{
\textbf{\textit{(1) Direct ST generation}}
} \\

GPT-5.5
  & 71.11 & 1.67
  & 75.58 & \textbf{7.05}
  & \textbf{18.06} & 65.19
  & 97.83 & \textbf{0.00}
  & 93.06 & 1.77
  & 27.87 & \textbf{32.24}
  & 37.88 \\

GPT-4o
  & 14.44 & 22.37
  & 8.91  & 42.06
  & 0.00  & \textbf{28.57}
  & 28.99 & 13.49
  & 1.04  & 21.11
  & 0.00  & 57.71
  & 92.75 \\

Claude Sonnet 5
  & 75.00 & 2.27
  & 75.91 & 25.50
  & 0.00 & 73.33
  & 88.41 & 0.72
  & 77.43 & 1.04
  & 4.37 & 51.72
  & 50.90 \\

Gemini 3.1 Pro
    & 78.33 & \textbf{0.00}
    & \textbf{80.86} & 18.09
    & 11.81 & 65.48
    & 94.20 & 1.45
    & \textbf{93.40} & 1.04
    & \textbf{35.25} & 51.95
    & 36.17 \\

  Gemini 3.8 Flash
  & \textbf{95.56} & \textbf{0.00}
  & 76.90 & 18.02
  & 4.86 & 72.73
  & \textbf{98.55} & \textbf{0.00}
  & 87.50 & \textbf{0.00}
  & 17.21 & 32.37
  & 32.10 \\

Qwen3.5-Plus
  & 60.56 & 6.71
  & 55.12 & 34.88
  & 0.00 & 57.14
  & 86.23 & 2.17
  & 46.18 & 24.11
  & 1.64 & 61.70
  & 49.79 \\

\midrule

\rowcolor{gray!15}
\multicolumn{14}{l}{
\textbf{\textit{(2) Generation-and-verification frameworks
(GPT-5.5 backbone)}}
} \\

Direct (GPT-5.5, matched)
  & 70.00 & 5.00
  & 75.25 & 5.21
  & \textbf{22.92} & 70.83
  & 93.48 & \textbf{0.00}
  & 85.42 & 5.56
  & 28.69 & 34.43
  & 39.61 \\

LLM4PLC
    & 75.00 & \textbf{0.00}
    & \textbf{77.23} & 4.95
    & 0.00 & \textbf{43.75}
    & 97.83 & \textbf{0.00}
    & 84.38 & 6.25
    & 23.77 & 48.36
    & 43.97 \\

Agents4PLC
  & 73.33 & 3.33
  & \textbf{77.23} & 10.89
  & 0.00 & 79.17
  & 84.78 & \textbf{0.00}
  & 70.83 & \textbf{0.00}
  & 16.39 & 27.87
  & 49.05 \\

  AutoPLC
    & \textbf{80.00} & \textbf{0.00}
    & 75.25 & 4.95
    & 12.50 & 75.00
    & \textbf{100.00} & \textbf{0.00}
    & 87.50 & 7.29
    & \textbf{31.15} & \textbf{25.41}
    & 41.44 \\

SemaPLC
    & 76.67 & \textbf{0.00}
    & 65.35 & \textbf{0.99}
    & 12.50 & 85.42
    & 93.48 & \textbf{0.00}
    & \textbf{94.79} & \textbf{0.00}
    & 9.84 & 39.34
    & 45.45 \\

\bottomrule
\end{tabular*}
\vspace{2pt}
\parbox{0.98\textwidth}{
\scriptsize\raggedright
Direct-model rows pool repeated runs; matched direct is the GPT-5.5
baseline used for framework comparison.
}
\end{table*}


\subsection{Main Results}

Table~\ref{tab:main_results} reports SR and SVR by suite and
difficulty together with ExecutionGap.

\textbf{Plant-level performance.}
Figure~\ref{fig:sr_by_difficulty} shows substantially lower SR
on Hard tasks than on Easy and Medium tasks across both
generation settings. When both suites are pooled by case count,
direct GPT-5.5 achieves 82.70\% and 84.09\% SR on Easy and
Medium tasks, respectively, compared with 25.10\% on Hard tasks.
Other models and frameworks show similar declines.
These comparisons describe differences between task groups
rather than isolating the effect of dependency scope.

\textbf{Completion and safety.}
Higher SR is not necessarily accompanied by lower SVR.
On Hard Material Handling tasks, Gemini 3.1 Pro achieves
higher SR than GPT-5.5 (35.25\% vs.\ 27.87\%), but also
higher SVR (51.95\% vs.\ 32.24\%). Thus, task completion
alone does not capture observed safety violations, while
a lower SVR does not necessarily indicate better completion.
These results support reporting \textit{Task Success} and
\textit{Safety Violation} separately.

\textbf{Verification and overhead.}
The adapted generation-and-verification workflows show
mixed changes across task groups. Relative to the matched
GPT-5.5 baseline, AutoPLC improves both SR and SVR on Hard
Material Handling tasks but worsens both on Hard Motion
Control tasks (Table~\ref{tab:main_results}).
On the shared 18-task subset, all evaluated frameworks use
more model calls, tokens, and elapsed time than direct
GPT-5.5 generation (Table~\ref{tab:generation_efficiency}).
LLM4PLC and AutoPLC each use a median of four calls while
revising the initial program for only 5 and 2 of the 18
tasks, respectively. Overall, higher measured generation
overhead is not accompanied by consistent plant-level
gains in this comparison. Paired SR comparisons and internal workflow-status outcomes are reported in Appendices~\ref{app:confidence_intervals}
  and~\ref{app:framework_results}, respectively.

\subsection{Analysis}
\label{sec:analysis}

\begin{wrapfigure}{r}{0.3\linewidth}
\vspace{-21pt}
\centering

\includegraphics[width=0.98\linewidth]{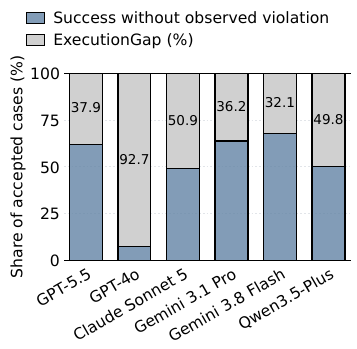}

\captionsetup{
  width=\linewidth,
  singlelinecheck=false,
}
\vspace{-5pt}

\caption{\textit{Execution Gap}}
\label{fig:execution_gap}

\vspace{-16pt}
\end{wrapfigure}

\subsubsection{Execution Gap}
\label{sec:execution_gap_analysis}

The reported \textit{Execution Gap} measures task failures or observed
safety violations among cases assigned to programs passing the
local submission-profile check, not external compilation or formal
verification. Table~\ref{tab:main_results} shows substantial gaps
in both generation settings; Figure~\ref{fig:execution_gap} shows
direct-generation results. Since acceptance rates vary, comparisons
require accepted/assigned counts. Gemini~3.8 Flash has lower
\textit{Execution Gap} than GPT-5.5 (32.10\% vs.\ 37.88\%), but fewer
accepted cases (1,187/1,419 vs.\ 1,399/1,419) and fewer assigned
cases achieving \textit{Task Success} without an observed \textit{Safety Violation}
(806/1,419 vs.\ 869/1,419). Lower \textit{Execution Gap} alone therefore
does not imply better overall outcomes.

\subsubsection{Failure Analysis}
\label{sec:failure_analysis}

\begin{figure*}[h]
\centering

\begin{minipage}[h]{0.66\textwidth}
\vspace{0pt}
\centering

\captionof{table}{Definitions of evaluator-recorded failure signatures.}
\label{tab:failure_signatures}

\scriptsize
\setlength{\tabcolsep}{2pt}
\renewcommand{\arraystretch}{1.06}

\begin{tabularx}{0.98\linewidth}{
  @{}
  >{\raggedright\arraybackslash}p{0.25\linewidth}
  >{\raggedright\arraybackslash}X
  @{}
}
\toprule
\textbf{Signature} & \textbf{Recorded manifestation} \\
\midrule

\colorbox{completionColor!40}{\textbf{\textit{Completion protocol}}}
& \\

Completion timing
& No valid completion is established by the required deadline \\

Completion validity
& Completion is declared before the required conditions are satisfied \\

Completion retention
& A required completed state or completion signal is not maintained \\

\midrule

\colorbox{safetyColor!40}{\textbf{\textit{Safety constraints}}}
& \\

Actuation constraints
& Actuation conflicts with another command or violates an interlock \\

Spatial constraints
& Execution violates a collision, clearance, or travel constraint \\

Resource constraints
& Occupancy, capacity, holding, or support requirements are violated \\

\midrule

\colorbox{behaviorColor!40}{\textbf{\textit{Required behavior}}}
& \\

Goal fulfillment
& The requested operation or required physical goal is not achieved \\

Process conformance
& Required order, route, identity, direction, or quantity is incorrect \\

\bottomrule
\end{tabularx}

\end{minipage}%
\hfill%
\begin{minipage}[h]{0.3\textwidth}
\vspace{0pt}
\centering

\includegraphics[width=0.98\linewidth]
{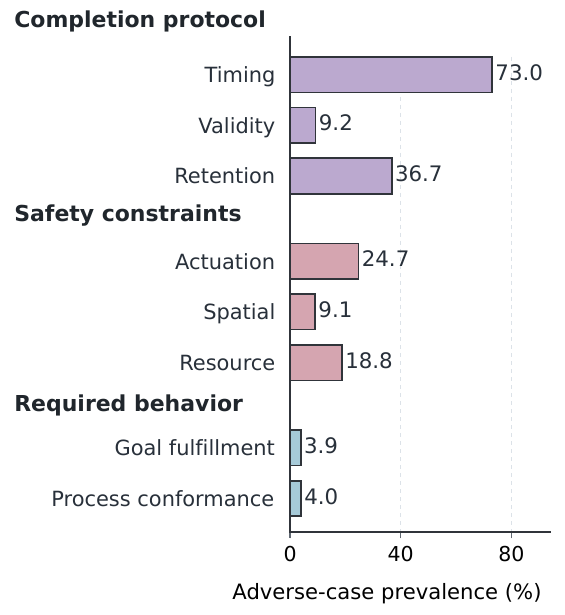}

\captionsetup{
  width=\linewidth,
  justification=raggedright,
  singlelinecheck=false,
  skip=3pt
}
\captionof{figure}{Failure-signature prevalence}
\label{fig:failure_signatures}

\end{minipage}

\end{figure*}

We examine how task failures and safety violations manifest
in cases entering plant execution after passing the common
upstream check. The analysis includes successful executions
with a violation but excludes generation and pre-execution
failures, which provide no plant observations.
Table~\ref{tab:failure_signatures} defines eight evaluator-recorded
signatures grouped into completion protocol, safety constraints,
and required behavior. The fixed mapping is multi-label:
cases may activate multiple signatures, so percentages need not
sum to \(100\%\).
Figure~\ref{fig:failure_signatures} reports prevalence
  across \(4{,}339\) adverse cases pooled over models and
  methods (\(3{,}615\) with failed-rule detail); matched
  direct is counted once.

Completion timing is the most frequent signature (\(73.0\%\)),
while completion retention affects \(36.7\%\) of adverse cases,
reflecting failure to maintain a required completed state or
completion signal.
In one \textit{Motion Control} case, a GPT-5.5 program asserts its
  completion output, but the plant state after that scan fails
  the required completion condition; the evaluator records
  premature completion.
Actuation and resource constraint violations affect \(24.7\%\)
and \(18.8\%\) of adverse cases, respectively.

The pooled percentages describe the composition of adverse cases,
not overall failure rates, differences across models or methods,
or root causes in the code.
Appendix~\ref{app:failure_analysis} provides the failure breakdown
by stage, disaggregated results, counts, and mappings from rules
to signatures.
\section{Conclusion}
\label{sec:conclusion-en}

We present PLCWorld as a common resource for reproducible
  closed-loop evaluation of LLM-generated ST programs, with
  100 tasks, 473 registered task--condition pairs, and separate
  \textit{Task Success} and \textit{Safety Violation} judgments.
  Reference and alternative programs provide evidence for
  task feasibility and acceptance of tested valid implementations,
  while controlled counterexamples assess targeted violation
  detection.
  Evaluations of six LLMs and four generation-and-verification
  frameworks demonstrate distinct completion and safety outcomes
  across task groups.


\subsection*{AI Use Statement}
\label{sec:ai-use-en}

Generative AI tools were used to assist with manuscript revision, including drafting and refining passages, and minor code edits. The LLMs evaluated in our experiments and their generation procedures are described in Section~\ref{sec:experiments-en}. The authors take responsibility for the final manuscript and code, including all AI-assisted content.

\subsection*{Ethics Statement}
\label{sec:ethics-en}

Errors in industrial control programs can damage equipment
or endanger people. PLCWorld evaluates generated ST programs
in simulation and reports \textit{Task Success} and
\textit{Safety Violation} separately.
Its safety metric covers specified constraints under evaluated
conditions and does not constitute safety certification.
Physical deployment requires independent engineering review
and equipment-specific validation.
The benchmark tasks instantiate abstracted industrial control
relations using newly defined goals, I/O mappings, and parameters,
without reproducing the source implementations.
Confidential source programs and engineering documents are
excluded from the planned public release.
Four practitioners reviewed the public task specifications
for industrial plausibility, information sufficiency, completion
and safety clarity, and difficulty appropriateness.
Reviewer names and employers are omitted; only broad experience
ranges and areas of expertise are reported.

\subsection*{Reproducibility Statement}
\label{sec:reproducibility-en}

Section~\ref{sec:construction-en} and
Appendix~\ref{app:benchmark_validation} describe benchmark
construction and validation.
Appendices~\ref{app:task_corpus}
and~\ref{app:plant_model_details} document task specifications
and plant models.
Section~\ref{sec:experiments-en} and
Appendix~\ref{app:experimental_details} provide the prompts,
model configurations, framework adaptations, evaluation
versions, and statistical and efficiency measurement procedures.
Our code, simulation environment, benchmark tasks, evaluator,
submission interface, and baseline implementations are
publicly available through the project website at
\url{https://yunji0516.github.io/PLCWorld/}.
Confidential industrial source materials are not included.

\bibliographystyle{iclr2027_conference}
\bibliography{references}

\clearpage
\appendix
\section*{Appendix Contents}

\begingroup
\makeatletter
\c@tocdepth=2\relax
\setlength{\parskip}{0pt}

\newcommand{\apptoclink}[2]{%
  \@ifpackageloaded{hyperref}{\hyperref[#1]{#2}}{#2}%
}

\newcommand{\apptocsection}[4]{%
  \l@section{\apptoclink{#3}{\numberline{#1}#2}}{#4}%
}
\newcommand{\apptocsubsection}[4]{%
  \l@subsection{\apptoclink{#3}{\numberline{#1}#2}}{#4}%
}
\makeatother

\apptocsection{A}{Limitations}
  {sec:limitations}{14}

\apptocsection{B}{Benchmark Contents and Task Specifications}
  {app:benchmark_contents}{14}
\apptocsubsection{B.1}{Benchmark Composition}
  {app:task_statistics}{14}
\apptocsubsection{B.2}{Public Task Specification Format}
  {app:task_spec_format}{15}
\apptocsubsection{B.3}{Standard Execution Conditions}
  {app:execution_conditions}{15}
\apptocsubsection{B.4}{Representative Task}
  {app:representative_tasks}{15}

\apptocsection{C}{Plant Model Details}
  {app:plant_model_details}{16}
\apptocsubsection{C.1}{Operational Device Families}
  {app:device_models}{17}
\apptocsubsection{C.2}{Transition Rules}
  {app:transition_rules}{19}
\apptocsubsection{C.3}{Workpiece and Transferable-Fixture Relations}
  {app:workpiece_relations}{19}
\apptocsubsection{C.4}{Modeling Scope}
  {app:modeling_scope}{19}

\apptocsection{D}{Industrial Grounding and Task Construction}
  {app:construction}{20}
\apptocsubsection{D.1}{Industrial Source Materials and Relation Abstraction}
  {app:construction:sources}{20}
\apptocsubsection{D.2}{Synthetic Task Instantiation}
  {app:construction:synthetic}{20}

\apptocsection{E}{Evaluation Rules and Metric Definitions}
  {app:evaluation_rules}{21}
\apptocsubsection{E.1}{Execution Records and Temporal Conditions}
  {app:evaluation_rules:events}{21}
\apptocsubsection{E.2}{Task Success and Safety Violation}
  {app:evaluation_rules:success}{21}
\apptocsubsection{E.3}{Errors, Termination, and Metric Eligibility}
  {app:evaluation_rules:errors}{21}

\apptocsection{F}{Benchmark and Runtime Validation}
  {app:benchmark_validation}{22}
\apptocsubsection{F.1}{Task Specification Validation}
  {app:expert_review}{22}
\apptocsubsection{F.2}{Evaluator Judgment Validation}
  {app:validation_judgment}{25}
\apptocsubsection{F.3}{Specification--Evaluator Alignment Validation}
  {app:validation_alignment}{25}
\apptocsubsection{F.4}{ST Runtime Validation}
  {app:runtime_validation}{25}

\apptocsection{G}{Experimental Setup and Reproducibility}
  {app:experimental_details}{26}
\apptocsubsection{G.1}{Benchmark Version and Evaluation Units}
  {app:evaluation_environment}{26}
\apptocsubsection{G.2}{Model Configurations and Input Protocol}
  {app:model_configurations}{26}
\apptocsubsection{G.3}{Framework Adaptations and Resource Budgets}
  {app:framework_adaptations}{28}
\apptocsubsection{G.4}{Acceptance Checks and Metric Accounting}
  {app:upstream_checks}{29}
\apptocsubsection{G.5}{Statistical Analysis and Efficiency Measurement}
  {app:reporting_details}{29}

\apptocsection{H}{Additional Experimental Results}
  {app:additional_results}{29}
\apptocsubsection{H.1}{Execution Coverage and Joint Outcomes}
  {app:execution_coverage}{30}
\apptocsubsection{H.2}{Confidence Intervals and Generation Variability}
  {app:confidence_intervals}{31}
\apptocsubsection{H.3}{Framework Evaluation Outcomes}
  {app:framework_results}{32}
\apptocsubsection{H.4}{Generation Efficiency}
  {app:framework_cost}{33}

\apptocsection{I}{Execution Gap and Failure Analysis}
  {app:failure_analysis}{34}
\apptocsubsection{I.1}{Common Upstream Check and Execution Gap Counts}
  {app:failure_analysis:execution_gap}{34}
\apptocsubsection{I.2}{Rule-to-Signature Mapping}
  {app:failure_analysis:taxonomy}{35}
\apptocsubsection{I.3}{Outcome and Signature Distributions}
  {app:failure_analysis:distributions}{35}
\apptocsubsection{I.4}{Representative Evaluation Outcomes}
  {app:failure_analysis:traces}{36}

\endgroup
\clearpage

\section{Limitations}
\label{sec:limitations}

Runtime comparisons cover selected constructs and programs,
with residual cross-runtime differences
(Appendix~\ref{app:runtime_validation}).
Broader runtime coverage and validation of plant responses
against measurements from physical equipment remain future work. PLCWorld's tasks are grounded in PLC
programs and engineering documentation from real-world manufacturing
equipment. Limited source coverage constrains generalization across
industrial settings, while source confidentiality restricts
independent inspection of this grounding. Evaluation covers specified
requirements under registered execution conditions, leaving unmodeled
hazards and untested executions outside its scope. Reported
\textit{Execution Gap} is conditioned on the common upstream check for
the supported ST profile and required I/O interface. Framework
comparisons reflect our PLCWorld adaptations under method-specific
resource budgets.

\section{Benchmark Contents and Task Specifications}
\label{app:benchmark_contents}
\label{app:task_corpus}

This appendix describes the composition of PLCWorld and the
model-facing task specifications used for evaluation. We report
statistics for the 100 benchmark tasks, document the information
provided to the evaluated models, describe the standard execution
conditions, and provide representative task objectives and plant
configurations.
These descriptive statistics are not used to assign task difficulty,
which is defined by the scope of control dependencies in
Section~\ref{sec:benchmark_tasks}.


\begin{table}[h]
\centering
\caption{Benchmark composition by suite and difficulty. I/O counts
all distinct addresses declared in each task's public specification,
including \texttt{BOOL}, \texttt{INT}, and \texttt{DINT} signals.
Device counts denote distinct operational device families per task
under the author-defined 17-family taxonomy.}
\label{tab:corpus_summary}
\small
\begin{tabular}{lrrrrrr}
\toprule
\textbf{Suite / Difficulty}
& \textbf{\#Tasks}
& \textbf{\#Cond.}
& \textbf{I/O avg.}
& \textbf{I/O range}
& \textbf{Device avg.}
& \textbf{Device range} \\
\midrule
Material / Easy   & 15 & 46  & 8.0   & 5--12   & 1.20 & 1--2 \\
Material / Medium & 20 & 96  & 17.4  & 11--28  & 2.85 & 1--7 \\
Material / Hard   & 15 & 122 & 57.5  & 37--75  & 5.13 & 2--9 \\
Motion / Easy     & 15 & 60  & 8.9   & 7--12   & 1.07 & 1--2 \\
Motion / Medium   & 20 & 101 & 20.1  & 12--29  & 2.10 & 1--4 \\
Motion / Hard     & 15 & 48  & 128.4 & 64--146 & 3.87 & 3--5 \\
\midrule
Total             & 100 & 473 & 37.9 & 5--146  & 2.68 & 1--9 \\
\bottomrule
\end{tabular}
\end{table}

\subsection{Benchmark Composition}
\label{app:task_statistics}

Table~\ref{tab:corpus_summary} summarizes the number of tasks,
standard execution conditions, public I/O addresses, and
operational device families in each suite--difficulty group.
Condition counts are taken from the frozen evaluation manifest.
I/O counts are computed from the distinct addresses declared
in each public task specification.

For composition statistics, operational devices are classified
using the author-defined taxonomy of 17 families documented in
Appendix~\ref{app:device_families}. Multiple instances of the same
family within a task are counted once. The resulting counts
characterize device-family diversity rather than the number of
physical devices or controlled axes. These composition statistics
are descriptive and are not used to assign difficulty.

\paragraph{Requirement-family distribution.}
Table~\ref{tab:relation_grid} provides the suite--difficulty
breakdown of the aggregate requirement-family counts reported in
Figure~\ref{fig:control_relations}. A family is assigned only when
the public task specification imposes the corresponding control
obligation on the generated program; device presence alone is
insufficient. Assignments are non-exclusive: each task is counted
at most once per family but may belong to multiple families.
These annotations characterize task content and are separate from
difficulty, which is defined by the scope of control dependencies.

\begin{table*}[t]
\centering
\caption{Distribution of control requirement families by suite and
difficulty. Counts are descriptive and non-exclusive; a task may
contribute to multiple rows.}
\label{tab:relation_grid}
\small
\setlength{\tabcolsep}{4pt}
\renewcommand{\arraystretch}{1.05}
\begin{tabular}{@{}lrrrrrrr@{}}
\toprule
& \multicolumn{3}{c}{\textbf{Material Handling}}
& \multicolumn{3}{c}{\textbf{Motion Control}}
& \\
\cmidrule(lr){2-4}\cmidrule(lr){5-7}
\textbf{Requirement family}
& \textbf{Easy}
& \textbf{Medium}
& \textbf{Hard}
& \textbf{Easy}
& \textbf{Medium}
& \textbf{Hard}
& \textbf{Total} \\
\midrule
Positioning and stabilization
& 8 & 19 & 14 & 14 & 20 & 15 & 90 \\
Synchronization and compensation
& 0 & 1 & 1 & 4 & 4 & 9 & 19 \\
Access and clearance
& 0 & 12 & 14 & 2 & 17 & 15 & 60 \\
Gripping and support
& 5 & 13 & 14 & 1 & 4 & 1 & 38 \\
Quantity and inventory
& 4 & 6 & 8 & 0 & 0 & 0 & 18 \\
Identification and routing
& 4 & 7 & 14 & 0 & 0 & 15 & 40 \\
Data and reference validity
& 4 & 7 & 10 & 2 & 3 & 3 & 29 \\
Occupancy and capacity
& 1 & 2 & 15 & 0 & 0 & 15 & 33 \\
Processing and separation
& 0 & 1 & 14 & 0 & 0 & 0 & 15 \\
\bottomrule
\end{tabular}
\end{table*}

\subsection{Public Task Specification Format}
\label{app:task_spec_format}

Each PLCWorld task is distributed as a model-facing specification
that defines the required external behavior without prescribing an
internal program structure. Within the supported ST profile,
a submission may choose its internal variables, branching logic,
timers, and state-machine organization, provided that its execution
satisfies the published plant-level requirements.

The specification provides the task objective, completion deadline,
evaluation endpoint or termination condition, public I/O interface
and signal semantics, initial plant state, permitted operating
conditions, device-response semantics, completion requirements,
and safety constraints.
Where relevant, it also identifies intermediate events and
precedence relations required for completion. Device descriptions
state how task-relevant plant state changes in response to outputs
and how the resulting state is exposed through sensor inputs.
Table~\ref{tab:task_spec_fields} summarizes the principal information
contained in a public task specification.

\begin{table*}[h]
\centering
\caption{Principal components of a PLCWorld public task specification.
The specification defines required observable behavior while allowing
implementation choices within the supported ST profile.}
\label{tab:task_spec_fields}
\small
\setlength{\tabcolsep}{5pt}
\renewcommand{\arraystretch}{1.08}
\begin{tabularx}{0.97\linewidth}{@{}p{3.2cm}XX@{}}
\toprule
\textbf{Component}
& \textbf{Information provided}
& \textbf{Role in evaluation} \\
\midrule

Task objective
& Required functional outcome and, where applicable, the requested
target, item, quantity, destination, or operating mode.
& Defines the plant-level result the submitted program must achieve. \\

Timing and completion
& PLC scan period, completion deadline, evaluation endpoint or
termination condition, and any required completion signal or
post-completion retention.
& Defines scan timing, completion validity, required result retention,
and when evaluation ends. \\

Public I/O interface
& Signal names, addresses, types, and input/output semantics.
& Defines the interface through which the ST program observes and
controls the task environment. \\

Initial plant state
& Initial device states, positions, workpiece or fixture relations,
and relevant request or process state.
& Defines the starting state for each execution. \\

Operating conditions
& Permitted ranges or alternatives for device responses, initial
conditions, and external inputs.
& Defines the range within which evaluation conditions may vary. \\

Device-response semantics
& Task-relevant effects of activating and deactivating outputs,
including timing and transition conditions.
& Defines how program outputs can change device, process, workpiece,
or fixture state. \\

Sensor semantics
& Meaning of feedback inputs such as position, operation completion, holding,
occupancy, validity, or process state.
& Defines what evidence the program can use to determine the effect
of its previous commands. \\

Required events and relations
& Intermediate events, precedence requirements, handover conditions,
or state dependencies required for valid completion.
& Distinguishes valid task execution from reaching a similar final
state through an invalid sequence. \\

Safety constraints
& Prohibited actions or states and task-specific interlock
requirements.
& Defines the conditions monitored for \textit{Safety Violation}. \\

\bottomrule
\end{tabularx}
\end{table*}

The model-facing input excludes reference ST programs and evaluator
source code. It provides the task requirements and permitted
operating ranges, but not the complete configurations of individual
registered execution conditions. Submitted programs are evaluated
against the published requirements without requiring the reference
program's internal control structure.

\subsection{Standard Execution Conditions}
\label{app:execution_conditions}

A task may be evaluated under multiple execution conditions.
Each condition instantiates concrete values for quantities already
allowed by the public task specification, while leaving the task
objective, public I/O interface, completion requirements, and safety
constraints unchanged. Across PLCWorld, the 100 tasks define
473 standard task--condition pairs
(Table~\ref{tab:corpus_summary}).

Execution conditions may vary the initial plant state, device
response or timing values, and external inputs such as requests or
start conditions when these quantities are variable in the task
specification. Sensor feedback is not supplied as a fixed trace:
during execution it is generated from the plant state reached under
the selected condition and the submitted program's preceding
outputs.

The standard conditions are frozen as part of the benchmark release
and are also used during benchmark construction and validation.

Table~\ref{tab:condition_scope} summarizes what remains fixed and
what may vary across conditions of the same task.

\begin{table}[h]
\centering
\caption{Relationship between a task specification and its execution
conditions. Condition-specific values remain within the operating
ranges published for the task.}
\label{tab:condition_scope}
\small
\begin{tabularx}{0.94\linewidth}{@{}lX@{}}
\toprule
\textbf{Fixed across conditions}
& Task objective; public I/O interface and semantics; device-response
rules; completion requirements and deadline rules; safety
constraints. \\
\midrule
\textbf{May vary by condition}
& Initial plant state, permitted device-response or timing values,
and task-defined external inputs or request parameters. \\
\bottomrule
\end{tabularx}
\end{table}

\subsection{Representative Task}
\label{app:representative_tasks}

\begin{figure}[h]
      \centering
      \includegraphics[width=\linewidth]{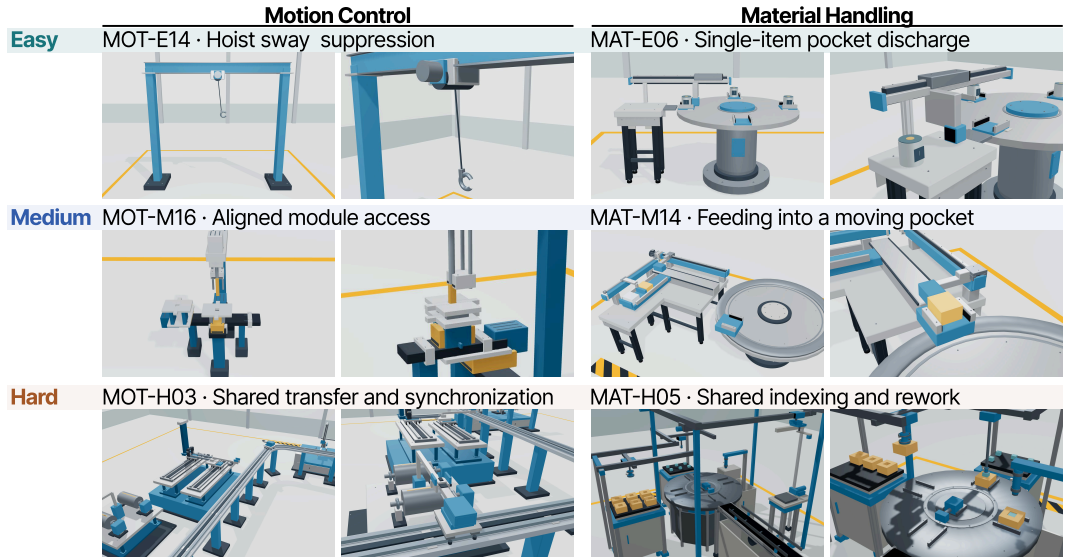}
      \caption{Plant configurations for the representative tasks in
  Table~\ref{tab:representative_tasks}, organized by task suite
  and difficulty level.}
      \label{fig:task_example}
  \end{figure}

Table~\ref{tab:representative_tasks} summarizes representative task
objectives across both suites and all three difficulty levels,
and Figure~\ref{fig:task_example} shows the corresponding plant
configurations. These examples illustrate the progression from a basic control function,
to dependencies between functions of one job, to coordination
between independent jobs. The complete public specification for
each task additionally contains the I/O interface, initial state,
operating conditions, device-response semantics, completion
requirements, and safety constraints described in
Appendix~\ref{app:task_spec_format}.

\begin{table*}[t]
  \centering
  \caption{Representative PLCWorld task objectives across suites
  and
  difficulty levels. Full model-facing specifications are
  distributed
  with the benchmark.}
  \label{tab:representative_tasks}
  \small
  \setlength{\tabcolsep}{5pt}
  \renewcommand{\arraystretch}{1.08}
  \begin{tabularx}{0.98\linewidth}{@{}llX@{}}
  \toprule
  \textbf{Task}
  & \textbf{Suite / Difficulty}
  & \textbf{Public objective} \\
  \midrule

  MOT-E14
  & Motion / Easy
  & Suppress the sway of an empty hoist hook and bring the trolley
  to rest at its current work position. \\

  MOT-M16
  & Motion / Medium
  & Position the table and carriage for tool access, maintain
  their
  alignment throughout the requested tool cycle, confirm
  completion
  of that cycle, and retract the carriage. \\

  MOT-H03
  & Motion / Hard
  & Coordinate two independent setup requests through lift
  stations,
  a shared shuttle, moving docking, and rotary synchronization,
  satisfying relative-motion constraints and returning each
  fixture. \\

  \midrule

  MAT-E06
  & Material / Easy
  & Discharge one product from the already aligned pocket and
  retract
  the pusher, while preserving the products in all other pockets.
  \\

  MAT-M14
  & Material / Medium
  & Prepare a single product, clear the feeder, and time its
  transfer
  into a continuously moving receiving pocket, while preserving
  the
  reserve product. \\

  MAT-H05
  & Material / Hard
  & Coordinate assembly, inspection, rework, and outfeed on a
  shared
  indexing table, preserving body--insert pairing and current
  inspection results, clearing all tools before indexing, and
  reinspecting reworked products before release. \\

  \bottomrule
  \end{tabularx}
  \end{table*}

The examples are illustrative rather than additional task
definitions. Evaluation always uses the complete frozen public
specification associated with each task ID.

\section{Plant Model Details}
\label{app:plant_model_details}

This appendix documents the executable plant abstractions used by
PLCWorld. The plant model is task-specific: each task instantiates
only the device, process, workpiece, fixture, and numerical states
needed to realize its public specification and evaluate task
completion and safety. The sections below describe the operational
device-family taxonomy used for corpus statistics, the transition
semantics used during execution, the representation of workpieces
and transferable fixtures, and the scope of the modeled physics.

\subsection{Operational Device Families}
\label{app:device_models}
\label{app:device_families}

Table~\ref{tab:device_catalog} classifies PLCWorld's operational
devices into 17 author-defined families. The same taxonomy underlies
the device-family statistics in Table~\ref{tab:corpus_summary},
counting each family at most once per task regardless of the number
of instances. This convention describes benchmark composition,
not an industry-standard taxonomy or the simulator's software
architecture.

The counts include devices that participate in task operation,
including externally driven and automatically operated devices.
Ordinary sensors, fixed supporting geometry, visual-only components,
inactive configuration branches, and internal components subsumed
by a higher-level device are excluded. Commands and feedback in the
table are illustrative semantic summaries; actual I/O names and
addresses, response parameters, and admissible transitions follow
each task's public specification.

{
\footnotesize
\setlength{\tabcolsep}{3pt}
\renewcommand{\arraystretch}{1.0}

\setlength{\LTcapwidth}{0.98\linewidth}

\begin{xltabular}{0.98\linewidth}{
    @{}
    >{\raggedright\arraybackslash}p{2.8cm}
    >{\raggedright\arraybackslash}p{3.0cm}
    >{\raggedright\arraybackslash}p{3.2cm}
    >{\raggedright\arraybackslash}X
    @{}
}

\caption{Author-defined operational device families in PLCWorld.
Commands and feedback are semantic summaries rather than universal
I/O names. Each family is counted at most once per task.}
\label{tab:device_catalog}\\

\toprule
\textbf{Family}
& \textbf{Typical command}
& \textbf{Typical feedback / state}
& \textbf{Modeled role} \\
\midrule
\endfirsthead

\multicolumn{4}{@{}l}{%
\textit{Table~\ref{tab:device_catalog} continued.}}\\
\toprule
\textbf{Family}
& \textbf{Typical command}
& \textbf{Typical feedback / state}
& \textbf{Modeled role} \\
\midrule
\endhead

\midrule
\multicolumn{4}{r@{}}{\textit{Continued on next page}}\\
\endfoot

\bottomrule
\endlastfoot

\texttt{linear\_axis}
& enable, extend/retract, target, rate
& position, velocity, endpoint, settled
& Positions a tool, fixture, counterweight, pusher, slide, or other
linearly moving mechanism. \\

\texttt{rotary\_axis}
& enable, index, target angle, rate
& angle, angular velocity, at-station, settled
& Orients or indexes a shaft, disk, table, tilt head, or slew
mechanism. \\

\texttt{lift\_hoist}
& raise/lower, target height, enable
& height, velocity, brake or settled state
& Changes the elevation of a supported load, tool, fixture, or
suspended item. \\

\texttt{transfer\_module}
& move, insert/retract, dock, couple
& position, inserted, docked, coupled, occupied
& Transfers a workpiece or fixture between supports, stations, or
process modules. \\

\texttt{manipulator}
& pose or taught-motion request, sequence
& at-pose, busy, motion complete
& Moves a tool or held item through an articulated or explicitly
identified Cartesian robot. \\

\texttt{mobile\_robot}
& motion enable, velocity, steering or target
& pose, heading, velocity, arrival
& Represents a mobile wheeled platform and its task-relevant motion. \\

\texttt{gripper}
& close, open, grip, release
& open, closed, holding
& Establishes or clears mechanical ownership of a workpiece. \\

\texttt{vacuum}
& vacuum on/off, acquire, vent
& vacuum present, holding, released
& Establishes or clears suction-based holding. \\

\texttt{clamp\_lock}
& clamp/unclamp, lock/unlock
& clamped, open, locked, holding
& Secures a workpiece, fixture, web, or docking relation in place. \\

\texttt{conveyor}
& run, stop, direction, speed
& occupied, position, boundary crossing
& Transports material on an explicitly modeled belt or powered roller
while maintaining support. \\

\texttt{fluid\_delivery}
& pump, fill, drain, valve selection
& level, delivered quantity, line or valve state
& Transfers a modeled liquid quantity between a supply, line, and
container. \\

\texttt{web\_handling}
& feed, wind, unwind, nip
& length, feed position, storage, reel state
& Supplies, stores, or takes up continuous web material. \\

\texttt{cutting\_press}
& cut, press, punch, return
& busy, returned, cut or press complete
& Separates, punches, or mechanically forms material through a
modeled operation. \\

\texttt{process\_module}
& start, cycle, machine, couple
& ready, busy, done, process result
& Performs a task-specific machining, assembly, coupling, or surface
operation. \\

\texttt{inspection}
& measure, inspect, weigh
& busy, result valid, measured value, identity
& Produces a measurement or classification result associated with a
workpiece. \\

\texttt{gate\_door}
& open, close
& open, closed, clear, busy
& Controls access or clearance to a station or protected operating
region. \\

\parbox[t]{2.2cm}{%
  \raggedright
  \texttt{packaging}\\[-1pt]
  \texttt{\_applicator}}
& apply, place, cap, tape
& busy, applied, complete
& Adds or applies a modeled packaging component such as a cap or
tape. \\

\end{xltabular}
}

\subsection{Transition Rules}
\label{app:transition_rules}

Device states evolve according to task-specific response dynamics,
timing ranges, and transition preconditions rather than being
established instantaneously by commands.
For example, a gripping command establishes holding only after the
closing response has completed while the workpiece is in an
admissible acquisition configuration. Depending on the task, an
inadmissible command may be blocked, have no physical effect, or
be recorded as a safety violation.

The evaluator assesses \textit{Task Success} from the required goal
plant state and event history, subject to the specified deadline
and hold conditions. When a completion signal is required, the
plant state after applying it for one scan must satisfy the
completion conditions. Command or completion bits alone are
insufficient.

\subsection{Workpiece and Transferable-Fixture Relations}
\label{app:workpiece_relations}

For tasks involving material or transferable fixtures, the plant
model maintains the subset of state properties required by the task.
These properties are task-specific rather than fields that must
appear in every plant model. They commonly include:

\begin{itemize}
    \item \textbf{Location}: the current position, station, slot,
    pocket, layer, or region occupied by a workpiece or fixture.

    \item \textbf{Owner / holder}: the tool, carrier, support,
    station, or process module that currently has custody of the
    item. Ownership changes only through a modeled acquisition,
    release, placement, or handover transition.

    \item \textbf{Support status}: whether the item is held,
    supported, in transfer, released onto a valid receiver, or
    unsupported when that distinction is relevant to completion
    or safety.

    \item \textbf{Identity and request association}: a persistent
    identifier, token, order, lot, or destination used when an item
    must remain associated with its originating request or result.

    \item \textbf{Quantity and arrangement}: task-specific counts,
    lengths, material amounts, slots, layers, or sequence positions
    used for batching, stacking, kitting, storage, or dispatch.

    \item \textbf{Process and inspection state}: task-specific
    machining, cutting, assembly, measurement, acceptance, or
    classification results associated with an item.

    \item \textbf{Transition history}: acquisition, release,
    handover, acceptance, processing, or acknowledgement events
    needed to distinguish a valid execution from the same final
    state reached through an invalid sequence.
\end{itemize}

\textit{Motion Control} tasks need not contain a conventional material
workpiece. In such tasks, the relevant transferred entity may
instead be a fixture, carrier, jig, or request-associated load, and
some tasks evaluate only device motion and feedback states.

\subsection{Modeling Scope}
\label{app:modeling_scope}

PLCWorld uses task-specific plant models combining discrete events
with task-relevant numerical state rather than a general-purpose
physics simulator. The plant models represent device responses,
timing, sensor updates, command preconditions, and workpiece or
fixture transitions required to evaluate task completion and safety.
Depending on the task, the modeled state may also include quantities
such as position, velocity, occupancy, material amount, mass,
balance relations, and collision or clearance geometry.

Contact forces, friction, deformation, process quality, and
multi-body dynamics are modeled only to the extent explicitly
represented through published state variables, transition rules,
geometric checks, or numerical relations. Each plant model is an
executable abstraction of its task specification, not a complete
physical model of the corresponding equipment.

\section{Industrial Grounding and Task Construction}
\label{app:construction}

\subsection{Industrial Source Materials and Relation Abstraction}
\label{app:construction:sources}

We cross-referenced PLC programs with the engineering documents
listed in Table~\ref{tab:industrial_source_roles} to interpret
commands, feedback signals, and device states. No single artifact
was treated as a complete task specification.

We abstracted control relations specifying an action, its observable
state or feedback, prerequisite device or workpiece states, and
states to maintain during subsequent operations. Source-specific
tags, vendor functions, numerical setpoints, and equipment layouts
were excluded from the public tasks. The source PLC programs and engineering documents are not included in the benchmark release.

\begin{table}[h]
\centering
\caption{Roles of the industrial source materials in task construction.}
\label{tab:industrial_source_roles}
\small
\begin{tabular}{p{0.23\linewidth}p{0.33\linewidth}p{0.34\linewidth}}
\toprule
\textbf{Source material} &
\textbf{Information consulted} &
\textbf{Use in task construction} \\
\midrule

PLC programs &
Actuation conditions, retained states, interlocks, and
request--completion logic &
Identification of command--feedback and sequencing relations \\

I/O lists and electrical drawings &
Signal direction, device association, and feedback source &
Interpretation of commanded and observed plant states \\

Equipment manuals &
Device operation, completion indication, and operating constraints &
Interpretation of task-relevant device and sensor behavior \\

Operating-sequence and engineering documents &
Operation order, readiness conditions, and handover context &
Identification of precedence, support, clearance, and
shared-resource relations \\

\bottomrule
\end{tabular}
\end{table}

\subsection{Synthetic Task Instantiation}
\label{app:construction:synthetic}

 For each task, we selected a functional objective informed by
  industrial equipment documentation and instantiated the associated
  control relations in a new synthetic plant configuration.
  We specified the devices and workpieces, initial plant state,
  public I/O interface and semantics, and plant-model parameters.

\begin{figure*}[t]
      \centering
      \includegraphics[width=\linewidth]{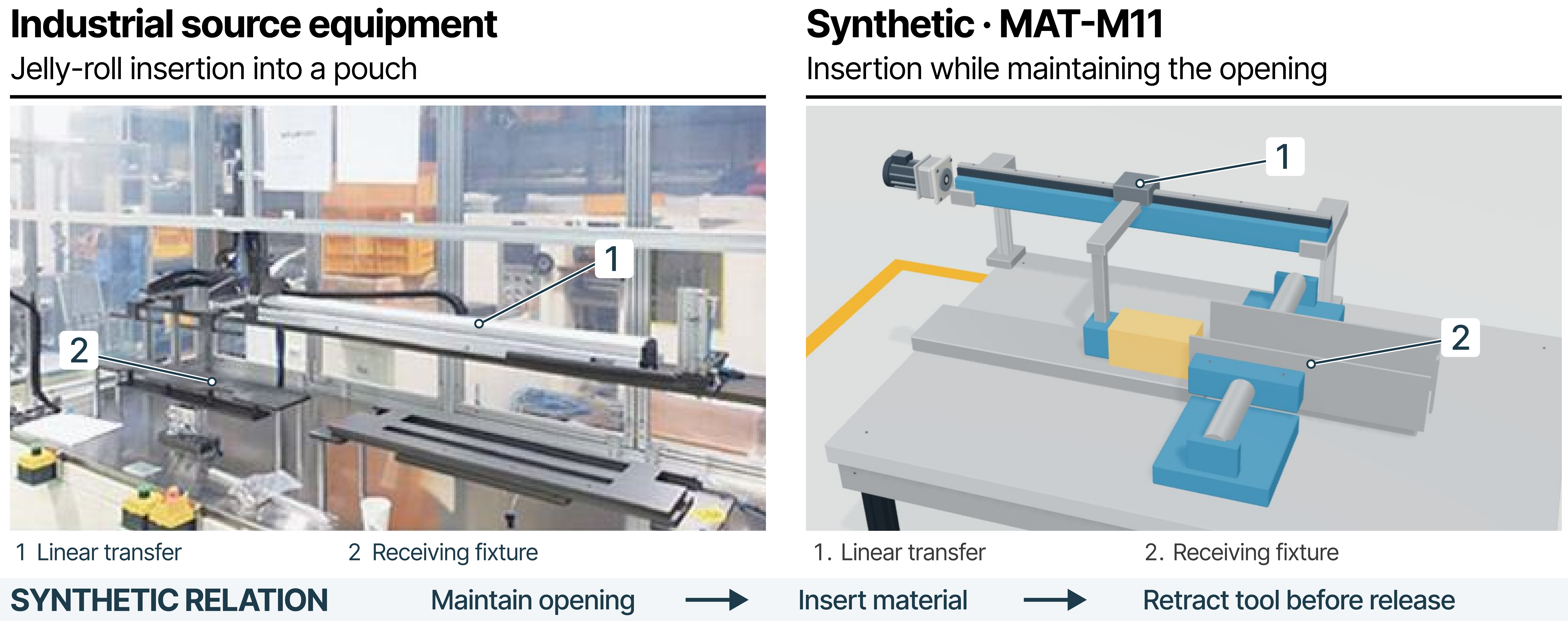}
      \caption{Industrial source equipment and synthetic PLCWorld task
MAT-M11. Matching numbers indicate corresponding device roles; the
bottom sequence summarizes the synthetic control relations.}
      \label{fig:industrial-task-correspondence}
  \end{figure*}

  Figure~\ref{fig:industrial-task-correspondence} compares
  industrial source equipment with a synthetic PLCWorld task.
  Matching annotations identify corresponding device roles,
  while the accompanying sequence summarizes the control relations
  specified for the synthetic task. This comparison illustrates
  task abstraction, rather than equivalence in physical layout,
  control implementation, or dynamics.

Each task specification defines the required event ordering
  and the plant states that must be maintained during subsequent
  operations. Completion requirements and safety constraints are
  expressed over plant states and events, without prescribing
  the program's internal implementation. Registered execution
  conditions assign concrete values within the task's permitted
  operating ranges (Appendix~\ref{app:execution_conditions}).

\section{Evaluation Rules and Metric Definitions}
\label{app:evaluation_rules}

\subsection{Execution Records and Temporal Conditions}
\label{app:evaluation_rules:events}

PLCWorld evaluates program--plant interaction using execution
records. In each completed scan, the program reads the current
input image and commits its outputs; the plant then updates its
device and workpiece states. The record contains the inputs,
committed outputs, updated plant state, and events. The evaluator
also derives task-defined events from recorded states and outputs.
Plant-level judgments do not depend on internal program variables
or presumed control strategy.

State conditions describe plant states and, where required,
committed outputs. Events mark occurrences such as acquisition,
arrival, or handover. Precedence relations specify the required
ordering of states or events; hold relations require a state
to persist over a specified interval. These conditions and
relations implement the public task requirements.
Completion claims are checked against the plant state after
the scan's outputs take effect and the event history through
that scan. Issuing a command or claim alone does not establish
the corresponding physical result.

\subsection{Task Success and Safety Violation}
\label{app:evaluation_rules:success}
\label{app:evaluation_rules:safety}

For a metric-eligible case \(e\), \textit{Task Success} \(U_e=1\) only
if all functional requirements are satisfied under its assigned
execution condition. These include the goal state, required
events and ordering, and any holding or retention requirements.
The goal state and any required valid completion declaration
must be established by the task-specific deadline; otherwise,
\(U_e=0\), even if observation continues.
A premature declaration does not establish completion;
its further consequences follow the task specification.
Loss of a required retained state or signal invalidates success.

For an eligible case entering plant execution, \textit{Safety Violation}
\(V_e=1\) if at least one specified safety constraint is breached
during observation. Monitoring covers the observed execution,
including any required period after a completion claim.
Once observed, a violation remains counted even if the plant
recovers. \textit{Task Success} and \textit{Safety Violation} are judged separately,
although some requirements affect both. Functional failure alone
is not a safety violation: success with a violation and failure
without an observed violation are both possible.
For interrupted executions, \(V_e=0\) means only that no violation
was observed before interruption.

\subsection{Errors, Termination, and Metric Eligibility}
\label{app:evaluation_rules:errors}

Cases with missing or unusable submissions, submission-profile
violations, parsing failures, unsupported constructs attributable
to the submitted program, or program-attributable execution errors
remain eligible for SR with \(U_e=0\).
Cases that never enter plant execution have no value of \(V_e\)
and are excluded from SVR. For program-attributable interruptions
after plant entry, the evaluator retains the observed execution
prefix and any recorded violation; \(U_e=0\).

Normal termination follows the task-specific evaluation endpoint,
which may be a deadline, a specified event, or the end of a
required post-declaration observation period.
A completion claim does not override this rule.
Records identify plant entry, completion of the required
observation, and any interruption point.

Infrastructure errors are excluded from all metric aggregation
and reported separately. They are not converted into task failures
or executions without an observed violation.

SR includes all metric-eligible assigned cases, whereas SVR
includes only metric-eligible cases that entered plant execution.
For an upstream check \(k\), the \textit{Execution Gap} denominator
contains only metric-eligible cases whose exact submitted ST
programs passed that check. Its numerator counts cases within
this set that exhibit task failure or an observed safety
violation, with each case counted once.
A metric with an empty denominator is reported as N/A.
The common upstream check used in the main comparison covers
the supported submission profile and required I/O interface,
not external compilation or formal verification.

\section{Benchmark and Runtime Validation}
\label{app:benchmark_validation}

This appendix details the validation procedures and results
summarized in Table~\ref{tab:validation_overview}, covering task
specifications, evaluator judgments, specification--evaluator
alignment, and ST runtime behavior.

\subsection{Task Specification Validation}
\label{app:expert_review}

Practitioner review complements executable checks by assessing whether
the assumptions and control relations described in a task are
reasonable for its intended industrial context%
~\citep{sargent2013verification}.
We conducted a practitioner review to examine whether PLCWorld's
public task specifications are industrially plausible and
sufficiently specified for implementation. The review focused
on properties that cannot be established solely through
reference-program execution or evaluator testing, including
industrial plausibility, information sufficiency, specification
clarity, and the practical appropriateness of the difficulty
levels.

\subsubsection{Practitioner Background}

The review involved four practitioners with professional
experience in PLC programming, equipment control, and industrial
automation. To preserve anonymity, we report broad experience
ranges and areas of expertise rather than names, employers,
or specific industrial projects.

Table~\ref{tab:expert_background} summarizes the reviewers'
backgrounds. Each reviewer had at least five years of professional
experience, with complementary expertise spanning PLC and control
engineering, automation system integration, commissioning, and
equipment control.

\begin{table}[h]
\centering
\small
\setlength{\tabcolsep}{3pt}
\renewcommand{\arraystretch}{1.08}

\caption{Background of practitioners participating in the practitioner
review. Names and employers are omitted to preserve anonymity.}
\label{tab:expert_background}

\begin{tabularx}{\linewidth}{
  @{}
  l
  >{\raggedright\arraybackslash}p{0.15\linewidth}
  >{\raggedright\arraybackslash}p{0.29\linewidth}
  >{\raggedright\arraybackslash}X
  @{}
}
\toprule
\textbf{Reviewer}
& \textbf{Experience}
& \textbf{Primary role}
& \textbf{Relevant expertise} \\
\midrule

E1
& 5--10 years
& PLC / Controls Engineer
& PLC programming, sequencing, interlocks \\

E2
& 5--10 years
& Automation / System Integration Engineer
& Material handling, system integration, industrial networks \\

E3
& 10+ years
& Commissioning Engineer
& Commissioning, I/O checkout, troubleshooting, motion control \\

E4
& 10+ years
& Equipment / Control Engineer
& Servo and pneumatic systems, equipment operation \\

\bottomrule
\end{tabularx}
\end{table}

\subsubsection{Review Scope and Procedure}

Reviewers examined the public, model-facing task specifications;
reference ST implementations and evaluator rules were not included
in the review materials. The 100 tasks were organized into six
suite--difficulty groups, with 15 Easy, 20 Medium, and 15 Hard
tasks per suite. Each reviewer examined all specifications in
each group, provided assessments for the group, and recorded
concerns about individual tasks using the corresponding task IDs.

The review covered four aspects:

\begin{enumerate}
    \item \textbf{Industrial plausibility}: whether the task
    objective, device configuration, and control relations are
    reasonable in an industrial automation setting;

    \item \textbf{Information sufficiency and clarity}: whether
    the specification provides sufficient information about
    device behavior, initial conditions, I/O semantics, and
    command--feedback relations to implement the required
    PLC logic;

    \item \textbf{Completion and safety clarity}: whether
    completion conditions, control dependencies, interlocks,
    and prohibited actions are stated clearly enough to
    determine correct behavior; and

    \item \textbf{Difficulty appropriateness}: whether the
    assigned level reflects the scope of control dependencies
    required by the task.
\end{enumerate}

For the first three aspects, the available ratings were
\textit{Appropriate}, \textit{Minor Revision Required},
\textit{Major Revision Required}, and \textit{Unable to Judge}.
For difficulty, reviewers assessed whether the assigned level
was consistent with the stated dependency scope and recorded
concerns about the assignment or its justification.

Reviewers could also record task-specific concerns involving
industrial plausibility, missing implementation information,
I/O or device behavior, completion or safety requirements,
implementation restrictions, or difficulty assignment.
A task could be associated with multiple issue types.
These comments were analyzed separately from the assessments
of each group.

\subsubsection{Review Results by Group}

Table~\ref{tab:expert_group_results} summarizes the assessments
of the six suite--difficulty groups. Across the 24 assessments
for each criterion, information sufficiency and completion and
safety clarity were rated \textit{Appropriate} in all cases.
Industrial plausibility received 16 \textit{Appropriate} and
8 \textit{Minor Revision Required} ratings, primarily reflecting
requests for clearer operational justification of selected
synthetic tasks. Difficulty feedback called for clearer
descriptions of dependency scope in two groups.
These ratings summarize each group as a whole; concerns about
individual specifications were recorded separately.

\begin{table*}[h]
\centering
\small
\setlength{\tabcolsep}{4pt}
\renewcommand{\arraystretch}{1.18}

\caption{Review assessments by suite and difficulty.}
\label{tab:expert_group_results}

\begin{tabularx}{\linewidth}{
  @{}
  ll
  *{3}{>{\centering\arraybackslash}X}
  >{\raggedright\arraybackslash}p{0.25\linewidth}
  @{}
}
\toprule

\textbf{Suite}
& \textbf{Difficulty}
& \textbf{Industrial\newline plausibility}
& \textbf{Information\newline sufficiency}
& \textbf{Completion and\newline safety clarity}
& \textbf{Difficulty\newline feedback} \\

\midrule

\multirow{3}{*}{\shortstack[l]{Motion\\Control}}
& Easy
& A: 3, MR: 1
& A: 4
& A: 4
& Appropriate \\

& Medium
& A: 2, MR: 2
& A: 4
& A: 4
& Clarify dependency scope \\

& Hard
& A: 2, MR: 2
& A: 4
& A: 4
& Appropriate \\

\cmidrule(lr){2-6}

\multirow{3}{*}{\shortstack[l]{Material\\Handling}}
& Easy
& A: 3, MR: 1
& A: 4
& A: 4
& Appropriate \\

& Medium
& A: 3, MR: 1
& A: 4
& A: 4
& Appropriate \\

& Hard
& A: 3, MR: 1
& A: 4
& A: 4
& Clarify dependency scope \\

\bottomrule
\end{tabularx}

\par\vspace{3pt}
\begin{minipage}{\linewidth}
\footnotesize
\raggedright
A: Appropriate; MR: Minor Revision Required.
Numbers indicate reviewer counts.
The difficulty column reports qualitative feedback rather
than reviewer counts.
\end{minipage}

\end{table*}

\subsubsection{Task-Specific Concerns and Revisions}

Reviewer comments identified opportunities to clarify operational
rationale, completion requirements, and the dependency scope
underlying difficulty assignments. Other comments concerned
wording that could impose an unnecessary action order when
alternative sequences satisfy the same completion and safety
requirements.
Table~\ref{tab:expert_task_revisions} summarizes the issue
categories and the corresponding specification revisions.

\begin{table}[h]
\centering
\small
\setlength{\tabcolsep}{3pt}
\renewcommand{\arraystretch}{1.10}

\caption{Task-specific concerns identified by reviewers and
corresponding revisions. Only categories with recorded concerns
are shown.}
\label{tab:expert_task_revisions}

\begin{tabularx}{\linewidth}{
  @{}
  >{\raggedright\arraybackslash}p{0.23\linewidth}
  >{\centering\arraybackslash}p{0.13\linewidth}
  >{\raggedright\arraybackslash}X
  @{}
}
\toprule
\textbf{Issue type}
& \textbf{\# Tasks}
& \textbf{Revision} \\
\midrule

Industrial plausibility
& 10
& Clarified the operational purpose and industrial rationale
of the affected tasks. \\

Completion condition
& 3
& Specified the required final plant state, completion signal,
and retention period where applicable. \\

Implementation restriction
& 4
& Made alternative action orders explicit where they satisfy
the same completion and safety requirements. \\

Difficulty rationale
& 7
& Clarified the functional dependencies and shared-resource
interactions supporting the existing difficulty assignments. \\

\bottomrule
\end{tabularx}

\par\vspace{3pt}
\begin{minipage}{\linewidth}
\footnotesize
\raggedright
Counts refer to distinct tasks within each category;
a task may appear in multiple rows.
Difficulty revisions clarified the rationale without changing
the assigned levels.
\end{minipage}

\end{table}

Table~\ref{tab:expert_revision_examples} presents representative
reviewer concerns, the specification revisions made in response,
and their effects on the stated requirements.

\begin{table*}[h]
\centering
\small
\setlength{\tabcolsep}{3pt}
\renewcommand{\arraystretch}{1.10}

\caption{Representative task specification revisions following
practitioner review.}
\label{tab:expert_revision_examples}

\begin{tabularx}{\linewidth}{
  @{}
  l
  >{\raggedright\arraybackslash}X
  >{\raggedright\arraybackslash}X
  >{\raggedright\arraybackslash}X
  @{}
}
\toprule
\textbf{Task}
& \textbf{Reviewer concern}
& \textbf{Revision}
& \textbf{Effect on specification} \\
\midrule

MOT-M02
& The prohibition before request acceptance did not explicitly
cover an enabled axis whose target equaled its current position.
& Specified that no commanded axis may be enabled before
acceptance of the \texttt{X0} request, even when its target
would produce no movement.
& Makes the start interlock explicitly apply to axis enabling,
not only to actual movement. \\

\addlinespace[3pt]

MOT-M17
& The return wording could imply a fixed axis order even when
another safe return sequence was possible.
& Specified when the feeder must remain retracted and explicitly
permitted overlapping return movements when the protected
geometry remains clear.
& Preserves the required outbound dependency while allowing
alternative safe return sequences. \\

\addlinespace[3pt]

MAT-H04
& Whether completion required physical receipt of the packages
and return of the handling devices was unclear.
& Specified receipt and acknowledgement of all four packages,
return of the empty shuttle, pickup head, and pusher, and
retention of the completed state for 0.5\,s.
& Makes physical receipt, device return, and retention
explicit parts of the completion requirements. \\

\addlinespace[3pt]

MAT-H13
& The effect of an exit request with an empty gripper and the
conditions for transporting a held bundle needed clarification.
& Clarified that an exit request with an empty gripper moves
nothing, while transferring a held bundle requires a free exit,
a released wrapping clamp, and separated film.
& Distinguishes an ineffective request from a valid bundle
transfer and makes the required interlocks explicit. \\

\bottomrule
\end{tabularx}
\end{table*}

\subsubsection{Feedback on Difficulty Levels}

Reviewers also assessed PLCWorld's dependency-based difficulty
definition independently of individual task wording.
The benchmark defines Easy tasks by dependencies contained within
one basic function, Medium tasks by dependencies between functions
of one job, and Hard tasks by dependencies between independent
jobs sharing resources, space, or plant state.

Three of the four reviewers considered this distinction generally
appropriate for representing differences in PLC control
complexity. The remaining reviewer considered it acceptable
provided that the explanations of dependency scope were clarified
in response to the feedback.

Reviewer comments identified seven tasks whose descriptions
required clearer justification of their assigned levels.
We revised these explanations to make the relevant functional
dependencies and coordination requirements explicit.
No tasks were reassigned: each suite retained 15 Easy,
20 Medium, and 15 Hard tasks, and the experiments used these
unchanged difficulty assignments.

We treat this assessment as evidence that the proposed levels
are meaningful from a practitioner perspective, rather than
as proof of an objective or universal ordering of PLC
programming difficulty.

\subsubsection{Specification Revision and Finalization}

We revised 18 task specifications to address the reviewer
concerns summarized above. The revisions clarified operational
rationale, completion requirements, permitted action sequences,
and the explanations supporting difficulty assignments.
Revised specifications were returned to the reviewers for
follow-up. The benchmark version used in the experiments
incorporated these revisions and was frozen after the
review-and-revision process.

\subsection{Evaluator Judgment Validation}
\label{app:validation_judgment}

\subsubsection{Feasibility and Valid Alternatives}
\label{app:validation_feasibility}

For each task, we run its reference ST program under every
registered execution condition. All 100 reference programs
achieve \textit{Task Success} without an observed \textit{Safety Violation} across
the 473 task--condition pairs. This establishes a feasible
implementation for each registered task and condition, but does
not make the reference program's internal control sequence part
of the acceptance criterion.

We also run 72 author-written alternative programs covering
67 tasks. These programs vary permitted control choices, including
preparation order or scheduling where the specification allows it.
All 378 corresponding executions achieve \textit{Task Success} without an
observed \textit{Safety Violation}. This test checks that the evaluator
accepts known valid implementations other than the reference; it
does not establish acceptance of every possible valid program.

\subsubsection{Targeted Violation Detection}
\label{app:validation_counterexamples}

We execute 542 controlled counterexample programs designed to
violate specified completion, ordering, retention, command, or
safety requirements. Each program has a prespecified evaluator
rule to test, and all 542 activate their designated rule in at
least one registered execution condition. These programs produce
739 executions across registered conditions, of which 723 are
rejected. The remaining 16 executions occur under conditions
in which those programs do not produce the targeted violation;
the same programs activate their designated rules under other
registered conditions. We therefore assess targeted detection
per program while retaining all execution outcomes in the
validation records.

\subsection{Specification--Evaluator Alignment Validation}
\label{app:validation_alignment}

We check the public task inputs against the selected execution
paths for all 100 tasks. The checks cover rendered input and
source hashes, scan configuration, I/O addresses, types and
directions, and submission entry points. All 100 tasks pass
these checks. A separate comparison of 461 published numerical
and operating-range items with the registered conditions finds
no unresolved mismatch.

These checks are complemented by a bidirectional semantic audit:
public obligations are compared with executable judgment
conditions, and judgment conditions are checked for a basis
in the public task specification. Identified discrepancies
are classified and corrected. Following corrections, we rerun
the full reference, alternative, and counterexample sets,
obtaining the results reported above. Affected frozen model
submissions are replayed without new model generation.
These tests provide evidence for the audited obligations and
registered conditions; they are not an exhaustive proof over
arbitrary ST programs or continuous plant states.

\subsection{ST Runtime Validation}
\label{app:runtime_validation}

\paragraph{Repeatability and diagnostic comparisons.}
Using the frozen evaluator, we rerun the reference suite in
two independent processes. Trace hashes and judgments match
across all 473 registered conditions, supporting repeatability
under the tested configuration.

Three diagnostic ST programs are compared between PLCWorld's
ST runtime and a pinned MATIEC/GCC
toolchain~\citep{beremiz2017matiec}, using identical input
images and virtual scan progression. All \(1{,}320\) observations
over \(360\) scans agree. The diagnostics exercise retained
state, \texttt{CASE}, loops, selected arithmetic,
\texttt{R\_TRIG}, and \texttt{TON}.
Siemens S7-PLCSIM Advanced~\citep{siemens2024plcsimapi} tests
examine \texttt{TON} scan boundaries; an identified discrepancy
was corrected and rechecked.

\paragraph{Comparison using generated programs.}
We use one admitted generation per task from each of six
direct-generation models, yielding 600 unchanged submissions.
Among those accepted by the pinned MATIEC compiler, we
compare 268 programs across 86 tasks under all registered
conditions (\(1{,}199\) paired cases).

Both runtimes execute identical source under the same
condition, with reset program memory and plant state and
the same frozen plant and evaluator. Each run evolves
independently through its own outputs and sensor feedback.

Complete scan-level records agree in \(1{,}162/1{,}199\) cases,
\textit{Task Success} judgments in \(1{,}195/1{,}199\), and
\textit{Safety Violation} judgments in all \(1{,}199\).
The 37 trace differences comprise 33 cases with matching
final judgments and four with differing \textit{Task Success}
judgments.

All four \textit{Task Success} differences involve a Qwen3.5-Plus
program for task MOT-E08 (Motion Control, Easy) that
calls the same \texttt{TON} instance twice per scan.
In these cases, PLCWorld records \textit{Task Success} false and
MATIEC/GCC true; neither records a \textit{Safety Violation}.
Another Qwen3.5-Plus program for task MOT-H09
(Motion Control, Hard) contributes three trace differences
with matching judgments.

The repeated-call difference is distinct from the corrected
Siemens \texttt{TON} scan-boundary discrepancy.
The comparisons cover the included programs and tested
execution paths, not full ST-profile equivalence or
independent validation of the shared plant and evaluator.

\section{Experimental Setup and Reproducibility}
\label{app:experimental_details}

This appendix documents the prompts, model configurations,
framework adaptations, and measurement protocols used in our experiments.

\subsection{Benchmark Version and Evaluation Units}
\label{app:evaluation_environment}

We use public model inputs v6 and evaluator v5.2.6.
Each repetition evaluates one ST program per task across
100 tasks and 473 task--condition cases. Programs remain
unchanged across their assigned conditions, with program
memory and plant state reset before each case. Only complete
repetitions without unresolved infrastructure errors enter
full-benchmark comparisons. Earlier-version artifacts require
documented replay confirming equivalent task-success and
safety outcomes. We retain versioned inputs, submitted programs,
and case-level evaluation records linked by source hashes.

\subsection{Model Configurations and Input Protocol}
\label{app:model_configurations}
\label{app:prompts}

\paragraph{Direct ST generation.}
Each direct-generation model receives the same system prompt
and task-specific user prompt in three independent single-turn
generations per task. Reference ST programs, evaluator source
code, and complete configurations of registered execution
conditions are withheld. Models have no access to tools,
execution feedback, or previous attempts. All outputs are
evaluated verbatim, without repair or best-of selection,
yielding 300 programs and 1,419 evaluation cases per completed
model.

\newtcolorbox{plcplantprompt}[1]{
  enhanced,
  breakable,
  colback=black!1,
  colframe=black!20,
  colbacktitle=black!6,
  coltitle=black,
  boxrule=0.4pt,
  arc=2pt,
  left=9pt, right=9pt, top=6pt, bottom=6pt,
  toptitle=4pt, bottomtitle=4pt,
  before skip=9pt, after skip=9pt,
  fonttitle=\sffamily\bfseries\small,
  fontupper=\small,
  before upper={\setlength{\parindent}{0pt}\setlength{\parskip}{3pt}},
  title={#1},
  title after break={#1\quad{\normalfont\footnotesize(continued)}}
}
\newenvironment{plcplantpromptitems}{%
  \begin{list}{\textbullet}{%
    \setlength{\leftmargin}{1.1em}%
    \setlength{\labelsep}{0.4em}%
    \setlength{\topsep}{2pt}%
    \setlength{\itemsep}{2pt}%
    \setlength{\parsep}{0pt}%
    \setlength{\partopsep}{0pt}%
  }%
}{\end{list}}
\newcommand{\plcpromptheading}[1]{%
  \par\smallskip{\sffamily\bfseries #1}\par\nobreak
}
\newcommand{\plcpromptomission}{%
  \par{\color{black!45}\texttt{[...]}}\par
}

\begin{plcplantprompt}{System prompt}
You write a PLC control program for the public task specification
supplied in the user message.

\plcpromptheading{Submission and I/O}
\begin{plcplantpromptitems}
\item Return exactly one complete IEC 61131-3 Structured Text
PROGRAM named Main, ending with \texttt{END\_PROGRAM}. Return source
only, without Markdown fences or explanatory prose.
\item Declare every listed I/O using its exact address identifier
and type. The task's single I/O table defines direction, access,
and meaning. Input signals are read-only: do not assign them, use
them as FOR counters, or connect function-block outputs to them.
Only declared output signals are writable.
\end{plcplantpromptitems}
\plcpromptomission

\plcpromptheading{Scan and completion semantics}
\begin{plcplantpromptitems}
\item At the start of a scan, external inputs and preceding device
feedback are sampled. ST executes on that input image. Its outputs
are then applied while device/workpiece states advance. Completion
is checked after that scan's state updates, using the actual state,
outputs and relevant history. Use the task's scan period, dynamics,
input scaling and operating limits; an output command is not proof
that its physical action has finished.
\end{plcplantpromptitems}
\plcpromptomission
\end{plcplantprompt}

\begin{plcplantprompt}{User prompt\hfill MAT-E01}
\textbf{MAT-E01 --- Product Acquisition with Gripper}
\plcpromptheading{Objective}
Acquire the single supported product with the parallel gripper.
\plcpromptomission

\plcpromptheading{I/O Interface}
{\setlength{\tabcolsep}{4pt}%
\renewcommand{\arraystretch}{1.12}%
\begin{tabularx}{\linewidth}{@{}lll>{\raggedright\arraybackslash}X@{}}
\toprule
\textbf{Address} & \textbf{Type} & \textbf{Direction} & \textbf{Meaning} \\
\midrule
\texttt{X10} & BOOL & input & Start request; one rising edge. \\
\texttt{D10} & INT & input & One product; D10=1. \\
\texttt{Y2} & BOOL & output & Completion claim, held until evaluation endpoint. \\
\texttt{Y20} & BOOL & output & grip: actuator command. \\
\texttt{X20} & BOOL & input & holding: physical observation. \\
\bottomrule
\end{tabularx}}

\plcpromptheading{Fixed I/O Declaration Skeleton}
\plcpromptomission
{\footnotesize
\begin{verbatim}
PROGRAM Main
VAR
    X10 : BOOL;
    D10 : INT;
    Y2 : BOOL;
    Y20 : BOOL;
    X20 : BOOL;
    (* Add internal variables here. *)
END_VAR
(* Implement the control logic here. *)
END_PROGRAM
\end{verbatim}
}
\plcpromptomission
\plcpromptheading{Safety Constraints}
\begin{plcplantpromptitems}
\item Do not initiate the requested operation before Start.
\end{plcplantpromptitems}
\plcpromptheading{Completion Conditions}
\begin{plcplantpromptitems}
\item The product belongs to the tool, the gripper is fully closed,
and grip remains ON.
\item Retain the result and Y2 through the remaining execution budget.
\end{plcplantpromptitems}
\plcpromptheading{Task-specific Operating Conditions}
\begin{plcplantpromptitems}
\item Gripper closing/holding and opening/release each respond in
0.10--0.20 s.
\item The evaluation window lasts 1600 scans.
\item Complete by scan 1550; at least 50 scans of result holding are checked.
\item The PLC scan period is 10 ms.
\end{plcplantpromptitems}
\end{plcplantprompt}
\noindent{\footnotesize\color{black!55}%
Prompt excerpts; \texttt{[...]} marks omitted text.
Formatting is adapted for readability.\par}

\begin{table}[!ht]
\centering
\small
\caption{Model configurations for the completed direct-generation
runs. ``Default'' indicates that the corresponding parameter was
omitted from the request. Output limits are reported in tokens.}
\label{tab:appendix_model_settings}
\setlength{\tabcolsep}{4pt}
\begin{tabularx}{0.98\linewidth}{@{}Xlll@{}}
\toprule
Model identifier & Reasoning / thinking & Temperature & \shortstack[l]{Max. output tokens} \\
\midrule
\texttt{gpt-5.5-2026-04-23} & Medium & Default & 32,768 \\
\texttt{gpt-4o-2024-08-06} & --- & Default & 8,192 \\
\texttt{claude-sonnet-5} & Default (adaptive) & Default & 65,536 \\
\texttt{gemini-3.1-pro-preview} & Default (high) & Default & 65,536 \\
\texttt{gemini-3.8-flash} & Medium & Default & 32,768 \\
\texttt{qwen/qwen3.5-plus-20260420} & Default (thinking) & Default & 65,536 \\
\bottomrule
\end{tabularx}
\end{table}

Qwen3.5-Plus uses synchronous OpenRouter requests at
concurrency eight; the other direct models use provider Batch APIs.
Token limits and reasoning-token accounting differ across
models, so inference compute is not matched. All frameworks
use the GPT-5.5 configuration in
Table~\ref{tab:appendix_model_settings}.

\subsection{Framework Adaptations and Resource Budgets}
\label{app:framework_adaptations}

We adapt LLM4PLC~\citep{fakih2024llm4plc},
Agents4PLC~\citep{liu2026agents4plc},
AutoPLC~\citep{yang2025autoplc}, and
SemaPLC~\citep{tu2026semaplc} to the same public task inputs
and plant evaluator. Table~\ref{tab:framework_resource_caps}
summarizes their workflows, checks, and per-task budgets.

\begin{table}[!ht]
\centering
\footnotesize
\caption{Workflows, internal checks, and per-task resource
limits configured for our PLCWorld adaptations.}
\label{tab:framework_resource_caps}
\setlength{\tabcolsep}{4pt}
\renewcommand{\arraystretch}{1.12}
\begin{tabularx}{0.98\linewidth}{@{}
>{\raggedright\arraybackslash}p{0.14\linewidth}
>{\raggedright\arraybackslash}X
>{\raggedright\arraybackslash}X
>{\raggedright\arraybackslash}X@{}}
\toprule
Framework & Generation workflow & Internal checks
& Configured limits per task \\
\midrule
LLM4PLC
& Control plan, ST generation, and feedback-based repair
& MATIEC compilation; nuXmv on a plan-level SMV model
& 16 model calls; 8 combined repairs; 700,000 total tokens \\
\addlinespace
Agents4PLC
& Retrieval, ranked plans, coding, debugging, and repair
& MATIEC compilation; nuXmv on a model translated from supported ST
& 24 model calls; 3 plans; 2 repairs per plan; 900,000 total tokens \\
\addlinespace
AutoPLC
& Planning, case/API retrieval, code generation, and repair
& Public submission-profile checks; CODESYS compilation
& 20 model calls; 3 compiler-feedback repairs; 800,000 total tokens \\
\addlinespace
SemaPLC
& Tool-assisted generation, checking, and revision
& Native project checks with candidate-specific verification records
& 60 model calls; 9 candidates; 12 native checks; 1,800\,s \\
\bottomrule
\end{tabularx}
\end{table}

\paragraph{Implementation adaptations.}
LLM4PLC targets the supported ST profile instead of SCL and
omits interactive human feedback. Agents4PLC is reconstructed
from the paper because the release we used lacks complete
orchestration and prompts. AutoPLC replaces hosted retrieval
with a deterministic local backend and omits LearnAgent,
which requires ground-truth programs. SemaPLC connects its
pinned public Bridge/Core implementation and native tools
to an isolated WSL worker through a local adapter.

\paragraph{Retrieval and verification details.}
Agents4PLC uses \texttt{text-embedding-ada-002} to retrieve
four entries from a fixed local corpus of public language
material and generic examples. AutoPLC retains 66 released
cases without modification after compatibility filtering.
BM25 retrieves five candidates, a model reranker selects
three, and API recommendation supplies at most fifteen
entries per prompt batch. PLCWorld tasks, reference programs,
and outcomes are not used to construct or filter this pool.

Agents4PLC derives verification properties from public
requirements without access to the selected plan or code.
Invalid properties are discarded without regeneration based
on code behavior; candidates with no valid properties remain
unverified. Passing these checks does not establish correctness
under plant execution.

\paragraph{Final submission and stopping rules.}
Full-benchmark evaluations use one run per task, with GPT-5.5
repetition~2 as the matched direct baseline. The retained final
candidate is evaluated despite failed internal checks, timeouts,
or budget exhaustion; cases without a usable submission count
as \textit{Task Success} failures. AutoPLC retains the last
candidate, and SemaPLC the latest complete candidate.
Documented SemaPLC infrastructure retries replace only affected
failed runs. Plant evaluation follows generation and internal
verification, and its outcomes are not used for candidate
selection or repair.


\subsection{Acceptance Checks and Metric Accounting}
\label{app:upstream_checks}

All metrics use pooled case counts and the error-handling
rules in Section~\ref{sec:evaluation}. \textit{Execution Gap} (EG) is
conditioned on the common upstream check for the supported
ST profile and required I/O interface, applied to the final
submitted ST program. We report EG with its eligible-case
denominator.


Framework-internal results are reported separately and apply
only to the checked program or model. A source-level pass
applies to the final submission only when its source matches
the checked version. Missing properties, unsupported
translations, timeouts, and tool errors do not count as
verification passes.

\subsection{Statistical Analysis and Efficiency Measurement}
\label{app:reporting_details}

\paragraph{Task-cluster bootstrap.}
We compute 95\% SR intervals using 10,000 stratified
task-cluster bootstrap replicates~\citep{field2007bootstrapping}
(NumPy default generator; seed \texttt{20260920}).
Tasks are sampled with replacement within the six
suite--difficulty groups, preserving group sizes and all
conditions and generation attempts for each sampled task.
SR is recomputed from pooled case counts, with intervals
given by the 2.5th and 97.5th percentiles. Paired SR
differences use identical task resamples. Intervals reflect
task resampling conditional on the generated programs
and are not adjusted for multiple comparisons.

\paragraph{Efficiency subset.}
We measure efficiency on 18 shared tasks, sampling three
without replacement from sorted task IDs in each of the six
suite--difficulty groups. Selection and execution order use
seeds \texttt{20260921} and \texttt{20260922}, respectively.
Direct models use the configurations in
Table~\ref{tab:appendix_model_settings}; frameworks use GPT-5.5.
All measurements use fresh synchronous runs at concurrency
one. Batch-job turnaround is not used as per-task latency.

Timing covers model calls, planning, retrieval, checking,
and repair, excluding shared initialization and plant
evaluation. Direct generation, LLM4PLC, Agents4PLC, and
AutoPLC use timers around their generation procedures.
SemaPLC uses process-start and final-result file timestamps,
which also capture local process and result-writing overhead.
These boundary differences limit fine-grained latency
comparisons.

Only methods with valid timing observations for all 18 tasks
enter the cost comparison. Internal verification failures,
timeouts, and budget exhaustion remain included; infrastructure
failures are excluded and reported separately.

For each task, we count LLM calls and sum their provider-reported
total tokens, excluding embedding calls and tokens. Revisions
count repairs or additional ST candidates after initial generation,
as recorded by each method. We report medians across the 18 tasks
for calls, revisions, tokens, and elapsed time, with the
interquartile range (IQR) for time.
Table~\ref{tab:generation_efficiency} also reports the number
of tasks with at least one revision. These measurements describe
observed API usage and latency rather than hardware-normalized
compute or monetary cost.

\paragraph{Additional operating conditions.}
\label{app:heldout_conditions}
The standard evaluation uses 473 registered task--condition pairs.
The reported results do not include a separate robustness
evaluation under additional held-out operating conditions.

\section{Additional Experimental Results}
\label{app:additional_results}

This appendix reports additional performance and efficiency
results using the protocols described in
Appendix~\ref{app:experimental_details}.

\subsection{Execution Coverage and Joint Outcomes}
\label{app:execution_coverage}
\label{app:joint_outcomes}

Table~\ref{tab:appendix_execution_coverage} reports case
counts at each evaluation stage, distinguishing failures
before plant entry from interrupted observations.

\begin{table}[h]
\centering
\normalsize
\setlength{\tabcolsep}{4pt}
\renewcommand{\arraystretch}{1.10}

\begin{threeparttable}
\caption{Case coverage under the common upstream check. Assigned, check-passed, and plant-entering counts are the SR, EG, and SVR denominators, respectively; these cohorts have no infrastructure exclusions.}
\label{tab:appendix_execution_coverage}
\begin{tabular}{@{}l*{5}{c}@{}}
\toprule
\multirow{3}{*}{\textbf{Model / Method}}
& \multicolumn{3}{c}{\textbf{Evaluation coverage}}
& \multicolumn{2}{c}{\textbf{Observation status}} \\
\cmidrule(lr){2-4}\cmidrule(lr){5-6}
& \textbf{Assigned}
& \shortstack{\textbf{Common check}\\[-1pt]\textbf{passed}}
& \shortstack{\textbf{Plant}\\[-1pt]\textbf{entered}}
& \shortstack{\textbf{Observation}\\[-1pt]\textbf{complete}}
& \textbf{Interrupted} \\
\midrule
\rowcolor{gray!15}
\multicolumn{6}{l}{\textbf{\textit{(1) Direct ST generation}}} \\
GPT-5.5 & 1419 & 1399 & 1399 & 1399 & 0 \\
GPT-4o & 1419 & 1062 & 1021 & 965 & 56 \\
Claude Sonnet 5 & 1419 & 1383 & 1383 & 1382 & 1 \\
Gemini 3.1 Pro & 1419 & 1316 & 1316 & 1316 & 0 \\
Gemini 3.8 Flash & 1419 & 1187 & 1176 & 1173 & 3 \\
Qwen3.5-Plus & 1419 & 960 & 946 & 936 & 10 \\
Direct (GPT-5.5, matched) & 473 & 462 & 462 & 462 & 0 \\
\midrule
\rowcolor{gray!15}
\multicolumn{6}{l}{\textbf{\textit{(2) Generation-and-verification frameworks (GPT-5.5)}}} \\
LLM4PLC & 473 & 473 & 473 & 473 & 0 \\
Agents4PLC & 473 & 473 & 473 & 473 & 0 \\
AutoPLC & 473 & 473 & 473 & 473 & 0 \\
SemaPLC & 473 & 473 & 473 & 473 & 0 \\
\bottomrule
\end{tabular}
\end{threeparttable}
\end{table}

Table~\ref{tab:appendix_joint_outcomes} cross-tabulates
\textit{Task Success} and observed \textit{Safety Violation}
among plant-entering cases. Task completion can coincide
with a safety violation, while task failure can occur
without an observed violation. For interrupted cases, the
absence of an observed violation does not establish safety
over the full evaluation endpoint.

\begin{table}[h]
\centering
\normalsize
\setlength{\tabcolsep}{6pt}
\renewcommand{\arraystretch}{1.10}
\begin{threeparttable}
\caption{Joint task and safety outcomes among plant-entering cases. The four outcome counts sum to the number of entered cases; pre-entry failures are excluded.}
\label{tab:appendix_joint_outcomes}
\begin{tabular}{@{}l*{5}{c}@{}}
\toprule
\multirow{3}{*}{\textbf{Model / Method}}
& \multicolumn{2}{c}{\textbf{\textit{Task Success}}}
& \multicolumn{2}{c}{\textbf{Task failure}}
& \multirow{3}{*}{\shortstack{\textbf{Plant}\\[-1pt]\textbf{entered}}} \\
\cmidrule(lr){2-3}\cmidrule(lr){4-5}
& \shortstack{\textbf{No observed}\\[-1pt]\textbf{violation}}
& \textbf{Violation}
& \shortstack{\textbf{No observed}\\[-1pt]\textbf{violation}}
& \textbf{Violation}
& \\
\midrule
\rowcolor{gray!15}
\multicolumn{6}{l}{\textbf{\textit{(1) Direct ST generation}}} \\
GPT-5.5 & 869 & 19 & 295 & 216 & 1399 \\
GPT-4o & 77 & 19 & 598 & 327 & 1021 \\
Claude Sonnet 5 & 679 & 47 & 341 & 316 & 1383 \\
Gemini 3.1 Pro & 840 & 91 & 190 & 195 & 1316 \\
Gemini 3.8 Flash & 806 & 57 & 193 & 120 & 1176 \\
Qwen3.5-Plus & 482 & 52 & 207 & 205 & 946 \\
Direct (GPT-5.5, matched) & 279 & 10 & 94 & 79 & 462 \\
\midrule
\rowcolor{gray!15}
\multicolumn{6}{l}{\textbf{\textit{(2) Generation-and-verification frameworks (GPT-5.5)}}} \\
LLM4PLC & 265 & 13 & 117 & 78 & 473 \\
Agents4PLC & 241 & 8 & 147 & 77 & 473 \\
AutoPLC & 277 & 21 & 117 & 58 & 473\\
SemaPLC & 258 & 6 & 125 & 84 & 473 \\

\bottomrule
\end{tabular}
\end{threeparttable}
\end{table}

\subsection{Confidence Intervals and Generation Variability}
\label{app:confidence_intervals}

Tables~\ref{tab:overall-sr-confidence-intervals}
and~\ref{tab:direct_sr_paired_ci} report pooled SR and
paired SR differences with 95\% confidence intervals,
using the task-cluster bootstrap described in
Appendix~\ref{app:reporting_details}. These intervals
quantify task-resampling uncertainty conditional on
the evaluated programs.

\begin{table}[h]
\centering
\normalsize
\setlength{\tabcolsep}{7pt}
\renewcommand{\arraystretch}{1.10}
\begin{threeparttable}
\caption{Pooled SR and task-cluster bootstrap 95\% confidence intervals. Direct models use three repetitions (1,419 cases); frameworks use one (473 cases).}
\label{tab:overall-sr-confidence-intervals}
\begin{tabular}{@{}l*{3}{c}@{}}
\toprule
\textbf{Model / Method} & \textbf{Successful / Assigned} & \textbf{SR (\%)} & \textbf{95\% CI (\%)} \\
\midrule
\rowcolor{gray!15}
\multicolumn{4}{l}{\textbf{\textit{(1) Direct ST generation}}} \\
GPT-5.5 & 888/1,419 & 62.58 & [55.28, 69.92] \\
GPT-4o & 96/1,419 & 6.77 & [4.38, 9.21] \\
Claude Sonnet 5 & 726/1,419 & 51.16 & [45.01, 57.09] \\
Gemini 3.1 Pro & 931/1,419 & 65.61 & [58.72, 72.52] \\
Gemini 3.8 Flash & 863/1,419 & 60.82 & [55.49, 66.24] \\
Qwen3.5-Plus & 534/1,419 & 37.63 & [31.80, 43.26] \\
\midrule
\rowcolor{gray!15}
\multicolumn{4}{l}{\textbf{\textit{(2) Generation-and-verification frameworks (GPT-5.5)}}} \\
LLM4PLC & 278/473 & 58.77 & [50.74, 67.11] \\
Agents4PLC & 249/473 & 52.64 & [44.66, 61.09] \\
AutoPLC & 298/473 & 63.00 & [54.51, 71.55] \\
SemaPLC & 264/473 & 55.81 & [48.53, 63.25] \\
\bottomrule
\end{tabular}
\end{threeparttable}
\end{table}

\begin{table}[h]
\centering
\normalsize
\setlength{\tabcolsep}{2pt}
\renewcommand{\arraystretch}{1.10}
\begin{threeparttable}
\caption{Paired differences in direct-generation SR, in percentage points (pp), using the same task clusters in both models. Intervals are pointwise and unadjusted.}
\label{tab:direct_sr_paired_ci}
\begin{tabular}{@{}lcc@{}}
\toprule
\textbf{Comparison} & \textbf{Difference (pp)} & \textbf{95\% CI (pp)} \\
\midrule
\rowcolor{gray!15}
\multicolumn{3}{l}{\textbf{\textit{(1) Direct ST generation}}} \\
GPT-5.5 $-$ GPT-4o & 55.81 & [48.61, 63.19] \\
GPT-5.5 $-$ Claude Sonnet 5 & 11.42 & [4.71, 18.67] \\
GPT-5.5 $-$ Gemini 3.1 Pro & $-$3.03 & [$-$9.87, 4.29] \\
GPT-5.5 $-$ Gemini 3.8 Flash & 1.76 & [$-$5.27, 8.98] \\
GPT-5.5 $-$ Qwen3.5-Plus & 24.95 & [17.59, 32.78] \\
GPT-4o $-$ Claude Sonnet 5 & $-$44.40 & [$-$50.11, $-$38.66] \\
GPT-4o $-$ Gemini 3.1 Pro & $-$58.84 & [$-$65.85, $-$52.11] \\
GPT-4o $-$ Gemini 3.8 Flash & $-$54.05 & [$-$59.63, $-$48.57] \\
GPT-4o $-$ Qwen3.5-Plus & $-$30.87 & [$-$35.97, $-$25.76] \\
Claude Sonnet 5 $-$ Gemini 3.1 Pro & $-$14.45 & [$-$21.13, $-$8.22] \\
Claude Sonnet 5 $-$ Gemini 3.8 Flash & $-$9.65 & [$-$15.80, $-$3.89] \\
Claude Sonnet 5 $-$ Qwen3.5-Plus & 13.53 & [8.48, 18.61] \\
Gemini 3.1 Pro $-$ Gemini 3.8 Flash & 4.79 & [$-$1.76, 10.99] \\
Gemini 3.1 Pro $-$ Qwen3.5-Plus & 27.98 & [21.30, 35.13] \\
Gemini 3.8 Flash $-$ Qwen3.5-Plus & 23.19 & [16.88, 29.73] \\
\bottomrule
\end{tabular}
\end{threeparttable}
\end{table}

Table~\ref{tab:framework_paired_ci} compares each framework's
  SR with both the matched direct run and the three-run direct
  baseline using paired task-cluster resampling. All intervals
  against the matched run include zero; against the three-run
  baseline, the intervals for Agents4PLC and SemaPLC are below zero.

\begin{table}[h]
  \centering
  \normalsize
  \setlength{\tabcolsep}{14pt}
  \renewcommand{\arraystretch}{1.10}
  \begin{threeparttable}
\caption{Framework-minus-direct GPT-5.5 differences in
  \textit{Task Success} rate (pp), with pointwise 95\% percentile CIs
  from 10,000 paired task-cluster bootstrap draws stratified
  by suite and difficulty. Draws retain all task conditions;
  the three-run baseline averages all direct repetitions.
  Intervals exclude framework rerun variability.}
  \label{tab:framework_paired_ci}
\begin{tabular*}{0.75\linewidth}{
@{\extracolsep{\fill}}lcc@{}
}
  \toprule
  \textbf{Framework}
  & \shortstack{\textbf{Matched direct run}\\\textbf{difference [95\% CI]}}
  & \shortstack{\textbf{Three-run direct mean}\\\textbf{difference [95\% CI]}} \\
  \midrule
  LLM4PLC
  & $-2.33$ [$-10.38$, $5.91$]
  & $-3.81$ [$-11.06$, $3.04$] \\
  Agents4PLC
  & $-8.46$ [$-17.87$, $1.06$]
  & $-9.94$ [$-17.93$, $-2.31$] \\
  AutoPLC
  & $1.90$ [$-6.29$, $10.35$]
  & $0.42$ [$-7.47$, $8.37$] \\
  SemaPLC
  & $-5.29$ [$-13.38$, $2.60$]
  & $-6.77$ [$-13.72$, $-0.21$] \\
  \bottomrule
  \end{tabular*}
  \end{threeparttable}
  \end{table}

Table~\ref{tab:appendix_generation_rounds} reports SR for
three independent generation repetitions, all included in
the pooled estimate. With only three repetitions, these
results provide a descriptive view of generation variability.

\begin{table}[h]
\centering
\normalsize
\setlength{\tabcolsep}{14pt}
\renewcommand{\arraystretch}{1.10}
\begin{threeparttable}
\caption{SR (\%) by generation repetition (100 programs and
473 evaluation cases each). Pooled SR includes all three
repetitions.}
\label{tab:appendix_generation_rounds}

\begin{tabular*}{0.75\linewidth}{
  @{\extracolsep{\fill}}l*{4}{c}@{}
}
\toprule
\multirow{2}{*}{\textbf{Model}}
& \multicolumn{3}{c}{\textbf{Repetition}}
& \multirow{2}{*}{\textbf{Pooled}} \\
\cmidrule(lr){2-4}
& \textbf{1} & \textbf{2} & \textbf{3} & \\
\midrule
\rowcolor{gray!15}
\multicolumn{5}{l}{\textbf{\textit{(1) Direct ST generation}}} \\
GPT-5.5          & 63.21 & 61.10 & 63.42 & 62.58 \\
GPT-4o           & 5.92  & 5.71  & 8.67  & 6.77  \\
Claude Sonnet 5  & 49.05 & 53.70 & 50.74 & 51.16 \\
Gemini 3.1 Pro   & 68.29 & 63.64 & 64.90 & 65.61 \\
Gemini 3.8 Flash & 63.21 & 58.56 & 60.68 & 60.82 \\
Qwen3.5-Plus     & 39.53 & 35.31 & 38.05 & 37.63 \\
\bottomrule
\end{tabular*}

\end{threeparttable}
\end{table}

\subsection{Framework Evaluation Outcomes}
\label{app:framework_results}

Table~\ref{tab:appendix_framework_results} reports overall
SR, SVR, and EG for the adapted frameworks and the matched
direct-generation baseline, all using GPT-5.5.
Results reflect each method under its configured resource
budget. EG is conditioned on the common upstream check.

\begin{table}[h]
\centering
\normalsize
\setlength{\tabcolsep}{6pt}
\renewcommand{\arraystretch}{1.10}
\begin{threeparttable}
\caption{Comparison with GPT-5.5 as the shared backbone. Each row uses one program per task and 473 assigned cases. EG cases give the numerator/denominator under the common upstream check.}
\label{tab:appendix_framework_results}
\begin{tabular}{@{}l*{4}{c}@{}}
\toprule
\textbf{Method} & \textbf{SR (\%)} & \textbf{SVR (\%)} & \textbf{EG (\%)} & \textbf{EG cases} \\
\midrule
\rowcolor{gray!15}
\multicolumn{5}{l}{\textbf{\textit{(1) Direct ST generation}}} \\
Direct (GPT-5.5, matched) & 61.10 & 19.26 & 39.61 & 183/462 \\
\midrule
\rowcolor{gray!15}
\multicolumn{5}{l}{\textbf{\textit{(2) Generation-and-verification frameworks (GPT-5.5)}}} \\
LLM4PLC & 58.77 & 19.24 & 43.97 & 208/473 \\
Agents4PLC & 52.64 & 17.97 & 49.05 & 232/473 \\
AutoPLC & 63.00 & 16.70 & 41.44 & 196/473 \\
SemaPLC & 55.81 & 19.03 & 45.45 & 215/473 \\
\bottomrule
\end{tabular}
\end{threeparttable}
\end{table}

\paragraph{Framework-internal checks and downstream outcomes.}
  Table~\ref{tab:framework_status_outcomes} links recorded
  workflow or check status to plant outcomes for the frozen
  final submissions. Program hashes match the evaluated ST
  submissions; neither generation nor plant evaluation was
  rerun. The rows partition each method's 100 tasks, but
  status labels have method-specific meanings.

  \begin{table}[h]
  \centering
\normalsize
  \setlength{\tabcolsep}{6pt}
  \renewcommand{\arraystretch}{1.10}
  \begin{threeparttable}
  \caption{Framework status and plant-level outcomes.
  SR is successful/assigned cases; SVR is cases with an
  observed safety violation/plant-entering cases.
  All 473 cases per method entered the plant. Status groups
  are descriptive and differ in task composition.}
  \label{tab:framework_status_outcomes}
  \begin{tabular}{@{}llrrr@{}}
  \toprule
  \textbf{Method} & \textbf{Recorded status}
  & \textbf{Tasks} & \textbf{SR} & \textbf{SVR} \\
  \midrule
  LLM4PLC & Plan formal accepted & 1 & 3/3 & 0/3 \\
          & Formal budget & 12 & 42/43 & 0/43 \\
          & Syntax budget & 1 & 4/4 & 0/4 \\
          & Timeout & 63 & 207/343 & 70/343 \\
          & Tool error & 23 & 22/80 & 21/80 \\
  \midrule
  Agents4PLC & Public properties verified & 11 & 37/40 & 0/40 \\
             & Plans exhausted: violated & 14 & 49/53 & 0/53 \\
             & Plans exhausted: vacuous & 6 & 16/26 & 0/26 \\
             & Timeout & 69 & 147/354 & 85/354 \\
  \midrule
  AutoPLC & CODESYS compiled & 94 & 289/452 & 62/452 \\
          & Verification budget exhausted & 6 & 9/21 & 17/21 \\
  \midrule
  SemaPLC & Native checks passed & 89 & 241/414 & 81/414 \\
          & Native compile failed & 4 & 3/26 & 0/26 \\
          & Native compile unchecked & 7 & 20/33 & 9/33 \\
  \bottomrule
\end{tabular}
\end{threeparttable}
\end{table}

  The checks do not share a common verification target.
  LLM4PLC's formal acceptance concerns an LLM-written
  plan-level SMV model, not a model derived from the final ST;
  its separate MATIEC check accepts 99/100 final programs.
  Agents4PLC checks public properties over its supported
  code-derived ST subset. AutoPLC records CODESYS~\citep{codesys} compilation,
  whereas SemaPLC records its native compile and runtime
  checks. A status without a recorded pass does not imply that
  no intermediate check was attempted. Because status groups
  have different task mixes, their outcome differences do not
  identify the effect of successful verification.

  Among compiler-accepted final programs, 208/469 LLM4PLC
  cases and 179/452 AutoPLC cases have task failure, an
  observed safety violation, or both. These adverse counts
  are not the complements of SR because successful cases
  can also violate safety constraints.

  The pinned MATIEC adapter rejects 31 of the 100 reference
programs: 19 involve conversion notation and 12 involve
mixed-type numeric expressions. Diagnostic copies using
explicit conversion functions and operand types compile
in all 31 cases. The rewrites change only conversion form and operand typing; we report compilation compatibility, not execution equivalence. We therefore use MATIEC for cross-runtime
comparison within the mutually accepted subset, rather
than as the common upstream check. The common
  submission-profile check and its \textit{Execution Gap} therefore
  remain the cross-method comparison in
  Table~\ref{tab:main_results}; the status breakdown gives
  method-specific evidence about final submissions.
  
\subsection{Generation Efficiency}
\label{app:framework_cost}

Table~\ref{tab:appendix_framework_cost} reports generation
efficiency for the six direct-generation models and four
GPT-5.5-based frameworks on the shared 18-task subset.
Measurement procedures and method-specific timing boundaries
are described in Appendix~\ref{app:reporting_details}.

\begin{table}[h]
\centering
\normalsize
\setlength{\tabcolsep}{6pt}
\renewcommand{\arraystretch}{1.10}
\begin{threeparttable}
\caption{Generation efficiency on the matched 18-task subset. Entries are medians; tokens are in thousands (k) and time in seconds. Time IQR gives the first and third quartiles, computed by inclusive linear interpolation.}
\label{tab:appendix_framework_cost}
\begin{tabular}{@{}l*{5}{c}@{}}
\toprule
\textbf{Method} & \textbf{Calls} & \textbf{Revisions} & \textbf{Tokens (k)} & \textbf{Time (s)} & \textbf{Time IQR (s)} \\
\midrule
\rowcolor{gray!15}
\multicolumn{6}{l}{\textbf{\textit{(1) Direct ST generation}}} \\
GPT-5.5 & 1 & 0 & 11.51 & 99.29 & [69.73, 124.38] \\
GPT-4o & 1 & 0 & 4.45 & 5.90 & [4.71, 9.42] \\
Claude Sonnet 5 & 1 & 0 & 17.00 & 122.32 & [49.91, 278.41] \\
Gemini 3.1 Pro & 1 & 0 & 19.17 & 118.09 & [49.06, 189.04] \\
Gemini 3.8 Flash & 1 & 0 & 25.80 & 65.54 & [20.06, 90.19] \\
Qwen3.5-Plus & 1 & 0 & 19.80 & 250.76 & [153.32, 324.74] \\
\midrule
\rowcolor{gray!15}
\multicolumn{6}{l}{\textbf{\textit{(2) Generation-and-verification frameworks (GPT-5.5)}}} \\
LLM4PLC & 4 & 0 & 48.28 & 321.13 & [254.40, 380.97] \\
Agents4PLC & 5 & 1 & 73.11 & 452.56 & [369.00, 760.94] \\
AutoPLC & 4 & 0 & 30.96 & 134.34 & [90.71, 150.86] \\
SemaPLC & 18 & 1 & 523.88 & 230.90 & [157.75, 275.87] \\
\bottomrule
\end{tabular}
\end{threeparttable}
\end{table}

These efficiency measurements use fresh runs on the 18-task
subset, separate from the 100-task benchmark evaluations.
Six of the 18 Gemini 3.8 Flash runs reached the output-token
limit configured for the efficiency evaluation. All 18
LLM4PLC efficiency runs ended in a timeout, tool error, or
budget exhaustion. The reported costs therefore include
incomplete verification workflows. These stopping outcomes
should not be interpreted as termination rates for the
full-benchmark evaluations.

\section{Execution Gap and Failure Analysis}
\label{app:failure_analysis}

\subsection{Common Upstream Check and Execution Gap Counts}
\label{app:failure_analysis:execution_gap}

The common upstream check used for Table~\ref{tab:main_results} is
PLCWorld's local submission-profile check, recorded as
\texttt{local\_submission\_profile}. It parses the submitted ST and
checks the required program structure, declared I/O types, read-only
inputs, and supported constructs. This check precedes runtime
initialization and plant execution; it is neither external native
compilation nor a framework-specific verification result.

The check result is linked to the same final ST program subsequently
evaluated by PLCWorld using a program hash. When a task-level submission passes,
all registered execution conditions assigned to that same program
enter the case-level acceptance set in Eq.~(\ref{eq:gap}). Thus, the
\textit{Execution Gap} denominator counts accepted task--condition cases,
not distinct programs. Subsequent initialization or execution errors
attributable to the submitted program count as task failures among
accepted cases. Table~\ref{tab:execution_gap_full} gives the numerator
and denominator of the values in the main table; accepted counts
should be considered when comparing conditional gaps.

\begin{table}[!htbp]
\centering
\captionsetup{font=normalsize,skip=4pt}

\normalsize
\setlength{\tabcolsep}{4pt}
\renewcommand{\arraystretch}{1.0}
\setlength{\aboverulesep}{1.5pt}
\setlength{\belowrulesep}{1.5pt}

\begin{threeparttable}

\caption{\textit{Execution Gap} under the common submission-profile check.
Accepted and adverse counts refer to metric-eligible task--condition
cases. Adverse cases have \textit{Task Success} failure, an observed \textit{Safety
Violation}, or both.}
\label{tab:execution_gap_full}

\begin{tabular*}{0.88\textwidth}{
  @{\extracolsep{\fill}}l*{4}{c}@{}
}
\toprule

\textbf{Model / method}
& \textbf{Assigned}
& \textbf{Accepted}
& \shortstack{\textbf{Adverse among}\\\textbf{accepted}}
& \shortstack{\textbf{Execution}\\\textbf{Gap (\%)}} \\

\midrule

\rowcolor{gray!15}
\multicolumn{5}{l}{
\textbf{\textit{(1) Direct ST generation}}
} \\

GPT-5.5
& 1,419 & 1,399 & 530 & 37.88 \\

GPT-4o
& 1,419 & 1,062 & 985 & 92.75 \\

Claude Sonnet 5
& 1,419 & 1,383 & 704 & 50.90 \\

Gemini 3.1 Pro
& 1,419 & 1,316 & 476 & 36.17 \\

Gemini 3.8 Flash
& 1,419 & 1,187 & 381 & 32.10 \\

Qwen3.5-Plus
& 1,419 & 960 & 478 & 49.79 \\

\midrule

\rowcolor{gray!15}
\multicolumn{5}{l}{
\textbf{\textit{(2) Generation-and-verification frameworks (GPT-5.5)}}
} \\

Direct (GPT-5.5, matched)
& 473 & 462 & 183 & 39.61 \\

LLM4PLC
& 473 & 473 & 208 & 43.97 \\

Agents4PLC
& 473 & 473 & 232 & 49.05 \\

AutoPLC
& 473 & 473 & 196 & 41.44 \\

SemaPLC
& 473 & 473 & 215 & 45.45 \\

\bottomrule
\end{tabular*}

\end{threeparttable}
\end{table}

Framework-internal checks are not pooled with this common check:
they may inspect different artifacts or apply different acceptance
criteria. The matched direct row uses a single generation per task
and is shown for framework comparison; its cases are not added again
to the pooled direct-generation results.

\subsection{Rule-to-Signature Mapping}
\label{app:failure_analysis:taxonomy}

Failure signatures describe evaluator-recorded manifestations,
     not source-code root causes. Signature analysis is restricted to
     cases that passed the common check, entered plant execution, and
     either failed the task or incurred an observed safety violation. A case may activate several
signatures. Cases that never enter the plant remain in the \textit{Execution Gap} and outcome analyses where applicable, but receive no plant-level
signature.

The fixed rule mapping first assigns each failed evaluator rule
  to one of ten detailed classes and then to one of the eight
  signatures reported in Section~\ref{sec:failure_analysis}.
  Table~\ref{tab:failure_taxonomy} shows this correspondence. The \texttt{support\_prohibited} class is
split by its recorded reason: support and material-integrity reasons
map to Resource constraints; other prohibited-operation reasons map
to Actuation constraints. This split uses the same fixed reason list
for every model and method.

\begin{table*}[t]
\centering

\normalsize
\setlength{\tabcolsep}{5pt}
\renewcommand{\arraystretch}{1.06}

\begin{threeparttable}

\caption{Mapping of ten rule-level classes to eight
evaluator-recorded failure signatures.
\texttt{support\_prohibited} is mapped according to
its recorded reason.}
\label{tab:failure_taxonomy}

\begin{tabularx}{0.88\textwidth}{
  @{}
  >{\raggedright\arraybackslash}p{0.21\textwidth}
  >{\raggedright\arraybackslash}p{0.34\textwidth}
  >{\raggedright\arraybackslash}X
  @{}
}
\toprule
\textbf{Signature}
& \textbf{Detailed rule class}
& \textbf{Grouping criterion} \\
\midrule

\colorbox{completionColor!40}{\textbf{\textit{Completion protocol}}}
& & \\

Completion timing
& \texttt{absent\_late}
& Missing or late completion \\

Completion validity
& \texttt{premature\_false}
& Invalid early declaration \\

Completion retention
& \texttt{persistence}
& Lost required state or signal \\

\midrule

\colorbox{safetyColor!40}{\textbf{\textit{Safety constraints}}}
& & \\

Actuation constraints
& \texttt{command\_conflict}; \texttt{support\_prohibited}
  for prohibited-operation reasons
& Command or interlock violation \\

Spatial constraints
& \texttt{collision\_clearance}
& Collision, clearance, or travel \\

Resource constraints
& \texttt{occupancy\_capacity}; \texttt{support\_prohibited}
  for support or material-integrity reasons
& Capacity, holding, or support \\

\midrule

\colorbox{behaviorColor!40}{\textbf{\textit{Required behavior}}}
& & \\

Goal fulfillment
& \texttt{request\_ack}, \texttt{physical\_goal}
& Unmet request or physical goal \\

Process conformance
& \texttt{sequence\_route}
& Order, route, identity, or quantity \\

\bottomrule
\end{tabularx}
\end{threeparttable}
\end{table*}

The mapping uses each failed rule's scope, identifier, and recorded
reason; the complete rule-identifier and reason inventory is retained
in the accompanying analysis artifact. An unseen rule stops the
mapping audit rather than entering a catch-all category. An adverse
case without failed-rule detail remains in the outcome count but
receives no inferred signature; in particular, it is not
automatically labeled as missing completion.

\subsection{Outcome and Signature Distributions}
\label{app:failure_analysis:distributions}

Tables~\ref{tab:appendix_execution_coverage}
and~\ref{tab:appendix_joint_outcomes} report execution coverage
and joint outcomes. After passing the common check,
41 GPT-4o, 11 Gemini 3.8 Flash, and 14 Qwen3.5-Plus cases
never enter plant execution. They count as failures in
\textit{Execution Gap} but have no safety judgment or
plant-level signature.

Table~\ref{tab:failure_family_incidence} reports signatures
by model and method among adverse plant-entered cases
passing the common check. Each case counts at most once
per signature, but labels overlap. The matched direct
baseline is counted only once in the pooled \(4{,}339\) cases.
Of these, 724 lack failed-rule detail and remain in the
denominator without assigned signatures. These distributions
describe recorded manifestations, not source-code root causes.

\begin{table*}[t]
\centering
\caption{Evaluator-observed signature incidence among adverse
plant-entered cases (\%). CT: completion timing; CV: completion
validity; CR: completion retention; GF: goal fulfillment; PC: process
conformance; AC: actuation constraints; SC: spatial constraints;
RC: resource constraints. ND is the percentage of adverse cases
without failed-rule detail. The eight signatures are multi-label;
their percentages need not sum to 100\%.}
\label{tab:failure_family_incidence}
\footnotesize
\setlength{\tabcolsep}{2.5pt}

\begin{tabular*}{\textwidth}{
  @{\extracolsep{\fill}}
  lrrrrrrrrrr
  @{}
}
\toprule
\textbf{Model / method}
& \textbf{Adverse \(n\)}
& \textbf{CT}
& \textbf{CV}
& \textbf{CR}
& \textbf{GF}
& \textbf{PC}
& \textbf{AC}
& \textbf{SC}
& \textbf{RC}
& \textbf{ND} \\
\midrule
\rowcolor{gray!15}
\multicolumn{11}{l}{\textbf{\emph{(1) Direct ST generation}}} \\

GPT-5.5
& 530 & 81.1 & 4.2 & 42.5 & 4.3 & 1.3 & 29.2 & 8.3 & 23.4 & 12.6 \\

GPT-4o
& 944 & 74.4 & 7.6 & 26.6 & 4.1 & 1.7 & 16.4 & 8.4 & 10.9 & 18.6 \\

Claude Sonnet 5
& 704 & 79.3 & 8.0 & 40.9 & 1.6 & 2.4 & 27.7 & 12.1 & 29.8 & 13.6 \\

Gemini 3.1 Pro
& 476 & 61.3 & 15.5 & 41.2 & 5.0 & 6.7 & 32.1 & 7.1 & 26.1 & 17.6 \\

Gemini 3.8 Flash
& 370 & 64.9 & 4.6 & 39.2 & 2.7 & 5.9 & 26.8 & 5.7 & 10.3 & 23.8 \\

Qwen3.5-Plus
& 464 & 63.8 & 23.7 & 28.7 & 6.7 & 4.7 & 18.1 & 13.6 & 19.4 & 20.9 \\

\midrule
\rowcolor{gray!15}
\multicolumn{11}{l}{\textbf{\emph{(2) Generation-and-verification frameworks (GPT-5.5)}}} \\

Direct (GPT-5.5, matched)
& 183 & 78.1 & 8.2 & 41.0 & 1.6 & 2.7 & 32.2 & 11.5 & 29.5 & 11.5 \\

LLM4PLC
& 208 & 77.9 & 2.4 & 38.9 & 1.4 & 8.7 & 25.0 & 5.8 & 14.9 & 13.9 \\

Agents4PLC
& 232 & 83.6 & 5.6 & 38.8 & 1.3 & 8.6 & 25.4 & 6.0 & 11.6 & 11.2 \\

AutoPLC
& 196 & 69.4 & 7.1 & 42.3 & 6.1 & 1.5 & 28.6 & 7.7 & 11.2 & 13.3 \\

SemaPLC
& 215 & 74.0 & 7.9 & 47.4 & 5.6 & 7.0 & 30.2 & 12.1 & 20.9 & 16.3 \\

\bottomrule
\end{tabular*}
\end{table*}

\subsection{Representative Evaluation Outcomes}
\label{app:failure_analysis:traces}

Table~\ref{tab:representative_failure_traces} illustrates distinct
outcomes and multi-label mapping. The examples were selected for
explanatory diversity rather than frequency.

\begin{table*}[h]
\centering
\caption{Representative case-level evaluator outcomes.
  The final column gives observed rule outcomes, not inferred
  source-code root causes. Here, r01 denotes the first generation
  run for each model or method.}
\label{tab:representative_failure_traces}
\normalsize
\setlength{\tabcolsep}{4pt}

\begin{tabular}{
  p{0.18\textwidth}
  p{0.15\textwidth}
  p{0.10\textwidth}
  p{0.10\textwidth}
  p{0.38\textwidth}
}
\toprule
\textbf{Model / run}
& \textbf{Task / \newline condition}
& \textbf{Task \newline Success}
& \textbf{Safety \newline Violation}
& \textbf{Recorded manifestation} \\
\midrule

GPT-5.5, r01
& MOT-E11 / 0
& No
& No
& Completion is asserted before the required post-transition plant
condition; \texttt{premature\_completion} is recorded. \\

GPT-5.5, r01
& MOT-H07 / 0
& Yes
& Yes
& The task reaches valid completion, but the execution also records
\texttt{head\_collision}. \\

LLM4PLC, r01
& MAT-H06 / 0
& No
& Yes
& \texttt{buffer\_overflow}, \texttt{wrong\_piece\_length},
\texttt{completion\_not\_held}, and \newline  \texttt{no\_completion}
activate several signatures in one case. \\

\bottomrule
\end{tabular}
\end{table*}

The stored case records identify the frozen program hash, registered
condition, evaluator version, failed rules, and execution-trace hash.
These observations establish the reported outcomes; identifying a
source-code root cause would additionally require inspection of the
submitted ST and scan-level trace.

\end{document}